%% file: colm2026_conference.tex
\documentclass{article} %
\usepackage[final]{colm2026_conference}

\usepackage{microtype}
\usepackage{hyperref}
\usepackage{url}
\usepackage{booktabs}
\usepackage{graphicx}
\usepackage{amsmath}
\usepackage{multirow}
\usepackage{adjustbox}
\usepackage{tcolorbox}
\usepackage{listings}
\usepackage{xcolor}
\usepackage{wrapfig}
\usepackage{booktabs}
\usepackage{colortbl}
\usepackage{subcaption}
\usepackage[T1]{fontenc}

\usepackage{lineno}

\definecolor{darkblue}{rgb}{0, 0, 0.5}
\hypersetup{colorlinks=true, citecolor=darkblue, linkcolor=darkblue, urlcolor=darkblue}

\definecolor{succbg}{HTML}{E8F0FE}
\definecolor{failtext}{HTML}{9AA0A6}
\newcommand{\win}[1]{\cellcolor{succbg}\textbf{#1}}
\newcommand{\fail}[1]{\textcolor{failtext}{#1}}

\title{The Depth Ceiling: On the Limits of Large Language Models in Discovering Latent Planning}

\author{%
  \mbox{} \And
  Yi Xu \\
  University of Cambridge\\
  \texttt{yx465@cam.ac.uk} \\
  \And \mbox{}
  \AND
  \mbox{} \And
  Philipp Jettkant\thanks{Equal contribution.}\\
  Imperial College London \\
  \texttt{p.jettkant@imperial.ac.uk} \\
  \And
  Laura Ruis$^{*}$ \\
  MIT \\
  \texttt{lruis@mit.edu}
  \And \mbox{}%
}

\begin{document}

\ifcolmsubmission
\linenumbers
\fi

\maketitle

\begin{abstract}
The viability of chain-of-thought (CoT) monitoring hinges on models being \emph{unable} to reason effectively in their latent representations. Yet little is known about the limits of such latent reasoning in LLMs. We test these limits by studying whether models can discover multi-step planning strategies without supervision on intermediate steps and execute them latently, within a single forward pass. Using graph path-finding tasks that precisely control the number of required latent planning steps, we uncover a striking limitation unresolved by massive scaling: tiny transformers trained from scratch discover strategies requiring up to three latent steps, fine-tuned GPT-4o and Qwen3-32B reach five, and GPT-5.4 attains seven under few-shot prompting. Although the maximum latent planning depth models can learn during training is five, the discovered strategy generalizes up to eight latent steps at test-time. This reveals a dissociation between the ability to \emph{discover} a latent strategy under final-answer supervision alone and the ability to \emph{execute} it once discovered. If similar limits hold more broadly, strategies requiring multiple coordinated latent planning steps may need to be explicitly taught or externalized, lending credence to CoT monitoring.
\end{abstract}

\section{Introduction}
\begin{figure*}[ht]
  \centering
  \includegraphics[width=\textwidth]{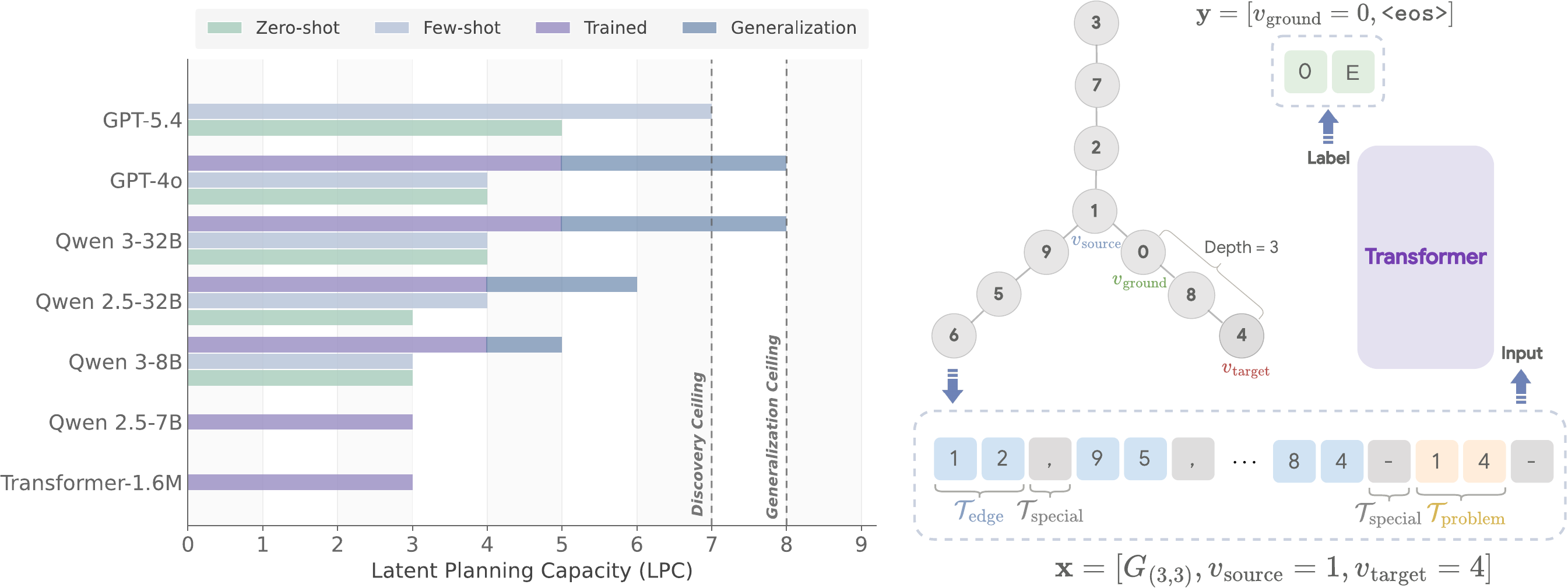}
  \vspace{-3mm}
    \caption{(Right) \textbf{We study latent planning using path-finding on star graphs.} Shown is an example of a star graph $G_{(3,3)}$ together with the corresponding input tokens and target label for training the transformer. (Left) \textbf{Latent planning capacity across models and evaluation settings.} Despite scaling from a 1.6M-parameter transformer to GPT-4o, the maximum LPC discovered during training increases by only two steps. However, LLMs can then generalize this strategy by a few more steps at test-time, up to 8 for the best models. As training is not publicly available for GPT-5.4, we report its few-shot LPC as a lower bound on the discovery ceiling, since fine-tuning strictly matches or exceeds few-shot performance for all other LLMs evaluated.}
    \label{fig:lpc_bar}
\end{figure*}

Chain-of-thought (CoT) reasoning is one of the main drivers of progress in language models. The discovery that prompting language models to reason step by step improves performance \citep{cot2022weietal, kojima2022large} led to large-scale efforts to optimize this ability. Today, frontier LLMs routinely produce long CoT traces \citep[inter alia]{openai2024openaio1card, Guo2025deepseek}. Beyond improving performance, externalized CoT provides an important lever for interpretability and oversight \citep{korbak2025chain}: overseers can read the reasoning trace to monitor what models are doing. Yet much less is known about what kinds of reasoning language models can carry out within a single forward pass, outside the chain of thought, and how these abilities evolve with scale. When intermediate reasoning steps are neither written out nor supervised, can models still discover and execute multi-step strategies internally, or do they depend on those extra tokens? Answering this question is an important step toward evaluating chain-of-thought monitoring as an oversight strategy.

We approach this question through the lens of planning, a setting in which success requires multi-step computation. Although planning does not capture all forms of latent reasoning, it offers a clean way to probe what models can discover and execute within a single forward pass, while precisely controlling reasoning depth and ruling out heuristic shortcuts. Specifically, we consider graph path-finding tasks in which the required number of latent planning steps can be controlled, and supervision is given only on task success. Models therefore receive a learning signal only when the full latent strategy is executed correctly. Recent work on such tasks has reached seemingly contradictory conclusions. \citet{bachmann2024pitfalls} report a complete failure of transformers to discover latent planning strategies when trained with next-token prediction, while \citet{saparov2025transformers} show that models can solve instances requiring many latent lookahead steps when trained with an implicit curriculum that includes shallow planning depths alongside harder instances, suggesting no architectural barrier to latent planning.

We argue that these results are reconciled by a key distinction between \textit{representational capacity} and \textit{strategy discovery}. Curriculum-based training shows that transformers can execute latent planning once training exposes them to the appropriate internal computation. Importantly, harder instances do not require qualitatively new strategies: increasing planning depth simply requires one additional iteration of the same latent computation. By including problems that require few planning steps alongside harder ones, such curricula allow models to bootstrap simpler strategies to more complex cases. What remains unclear is whether models can \emph{discover} such strategies when trained only on tasks requiring a single fixed planning depth, with supervision provided only on the final goal. In this setting, solving the task requires planning, but training provides no supervision over the intermediate steps needed to discover the strategy.

Our experimental setting uses graph path-finding on star graphs with a central source node and several equal-length branches (Figure~\ref{fig:lpc_bar}, right). In this setting, any above-random performance necessarily reflects multi-step latent planning. Crucially, models receive a learning signal only when all latent planning steps are executed correctly, and each model is trained only on problems requiring a single fixed depth, ruling out bootstrapping across depths. Success therefore requires discovering a complete internal computation under an all-or-nothing training signal. Across scales, from small transformers trained from scratch to pretrained LLMs, we find that models can discover deeper latent planning strategies than suggested by prior work \citep{bachmann2024pitfalls}, but remain surprisingly limited relative to curriculum-based results, which show near-perfect latent planning for at least 15 steps within a single forward pass, even in small models trained from scratch \citep{saparov2025transformers}. We report the following findings:
\begin{enumerate}
    \item Contrary to prior literature reporting a complete failure in our setting \citep{bachmann2024pitfalls}, a small transformer trained from scratch with next-token prediction can discover latent planning strategies requiring up to three internal lookahead steps. However, discovery fails abruptly at four steps and beyond. Increasing model depth or the number of attention heads does not overcome this limitation (see Appendix~\ref{appendix:hyperparameters}).
    
    \item Scaling up to frontier language models primarily improves the ability to handle greater planning breadth (more branches) rather than greater planning depth. While the from-scratch transformer fails at high branch factors, fine-tuned LLMs solve these configurations easily. Yet across 7B--32B open-weight models and GPT-4o, fine-tuning increases discovered planning depth from three steps to only four or five, with failure beyond that. Although we cannot fine-tune the most capable models out there today, we evaluate GPT-5.4 few-shot and find it achieves a maximum latent planning depth of seven steps.
    
    \item Strikingly, LLMs trained directly on six-, seven-, or eight-step problems never rise above random guessing, yet models trained on five-step problems discover a strategy that generalizes successfully to those same depths at test time. This leads to a slightly higher `generalization ceiling' at eight steps than the `discovery ceiling' at seven steps (Figure \ref{fig:lpc_bar}).

    \item Attention analysis on the transformer trained from scratch suggests that successful models learn a backtracking strategy that concentrates attention along the target-to-source path. Consistent with this, when fine-tuned LLMs are evaluated on depths beyond their training configuration, errors are predominantly on-path. In other words, models reliably identify the correct branch but fail to carry the backtracking process through all required planning steps, further underscoring the depth ceiling in generalization.
    
    \item Finally, in a control setting where models are allowed to externalize their reasoning by training with a backtracking strategy in the chain of thought, they successfully solve graphs requiring twenty lookahead steps (the max we attempt) with minimal effort, converging in only $20$ training updates. Combined with the generalization results above, this suggests that the primary bottleneck lies in discovering latent planning strategies under sparse supervision, rather than in the intrinsic difficulty of the task itself.

\end{enumerate}

Our findings reveal a persistent ceiling on the ability of frontier language models to discover latent planning strategies and execute them within a single forward pass, even in a simple, graph-search setting where deeper planning requires iterating the same computation. This is particularly striking because prior work shows that small transformers trained from scratch can learn much deeper latent planning when training on shallow and deep problems lets them build up the strategy step by step. Most surprisingly, scaling from a small eight-layer single-attention-head transformer to GPT-4o increases the depth of discoverable latent planning by only two steps. Although we cannot fine-tune the most capable frontier models available today, few-shot evaluations closely track fine-tuned ceilings in the models we can study, making them a useful proxy; under this evaluation, GPT-5.4 reaches only seven latent steps, suggesting that scale alone does not substantially improve latent strategy discovery.

While these results do not imply that deeper latent planning is impossible, they indicate that gradient-based learning with supervision only on task success provides weak training signals for discovering such strategies. This echoes broader evidence that latent reasoning abilities in LLMs are limited: models struggle with tasks requiring implicit multi-step inference \citep{berglund2023takencontextmeasuringsituational, balesni2025lessonsstudyingtwohoplatent}. Although the simple planning problems we study do not exhaust the space of latent reasoning, they shed light on the latent reasoning abilities of language models in a setting where reasoning depth can be precisely controlled. If latent reasoning is fundamentally shallow across reasoning domains, models may be forced to externalize the reasoning required for complex tasks, strengthening the case for chain-of-thought monitoring as a viable oversight strategy. Whether similar depth ceilings govern latent strategy discovery beyond planning represents an important line of future work that can help establish trust in CoT monitoring.

\input{secs/01_methods}

\input{secs/02_discussions}

\section{Related Work}

\paragraph{Implicit Planning}
Whether transformers can autonomously discover planning strategies under standard next-token prediction remains a fundamental open question. Although recent methods confirm that transformers can perform latent reasoning, these approaches typically rely on explicit training scaffolding, architectural modifications, or auxiliary supervision.
For instance, Coconut \citep{hao2024training} and ICoT \citep{deng2023implicit,deng2024explicit} progressively internalize full CoT into the latent space, while CODI \citep{hao2024training} distills hidden states from a teacher model with full CoT. Related work also proposes training signals tailored to the task structure, such as predicting latent states \citep{teoh2025nextlatentpredictiontransformerslearn} or inserting illegible intermediate tokens in place of explicit reasoning traces \citep{bachmann2024pitfalls}. While such methods can improve performance in controlled domains, they rely on prior knowledge about the computation to be learned, such as where latent reasoning steps should occur or how many such steps are required, and therefore do not address whether useful latent strategies emerge autonomously under standard large-scale training. They are also not representative of current frontier training practice, which overwhelmingly optimizes either standard next-token prediction or explicit chain-of-thought reasoning rather than task-specific latent-reasoning objectives.
This leaves open the question of whether latent planning can emerge under standard training without such scaffolding.
\cite{bachmann2024pitfalls} conclude that standard next-token prediction completely fails to induce true latent planning, causing models to exploit superficial greedy shortcuts (the \emph{Clever Hans cheat}), though their analysis does not examine shallow planning depths where success may still be possible. Conversely, \cite{saparov2025transformers} show that transformers can implicitly search graphs, but their training curriculum artificially guides strategy discovery by exposing intermediate planning depths, thereby obviating autonomous discovery. We address this question by demonstrating that when trained solely on sparse, outcome-based signals not admitting greedy shortcuts, models under standard next-token prediction can indeed autonomously discover and generalize implicit planning strategies. However, we reveal that this autonomous emergence is strictly bounded by a fundamental depth ceiling.

\paragraph{Implications for Chain-of-thought Monitoring}
Our findings have direct implications for the safety of CoT monitoring, which audits the explicit reasoning traces of LLMs to detect malicious or deceptive behavior \citep{korbak2025chain, pmlr-v235-greenblatt24a, baker2025monitoring}. The depth ceiling we identify suggests that current LLMs fundamentally require externalized reasoning for complex multi-step tasks, providing a safety margin for continued reliance on CoT monitoring. We discuss this connection and the broader related work on CoT monitoring in detail in Appendix~\ref{app:cot_monitoring}.

\section{Limitations and Future Work}
Our study focuses on star graphs, a highly symmetric structure lacking local heuristic cues. While real-world reasoning tasks (e.g., theorem proving and code generation) often contain denser signals or local heuristics that might partially bypass the discovery bottleneck, star graphs represent a lower bound on difficulty: they are the cleanest case with no heuristic shortcuts, and if models cannot discover latent strategies even here, that is informative about the fundamental limits of gradient-based learning under sparse supervision. Indeed, independent work on other domains has found similar limitations on latent multi-step reasoning in LLMs \citep{berglund2023takencontextmeasuringsituational, balesni2025lessonsstudyingtwohoplatent}, suggesting the depth ceiling may not be specific to planning. Whether this bottleneck persists in more diverse reasoning tasks remains an important open question.

A natural question is whether this limitation disappears in the strongest models available today. We cannot test this directly, since frontier closed-weight models are not accessible for fine-tuning. However, the pattern appears robust across the regimes we can test. Among the models we do fine-tune, zero- and few-shot evaluation reaches similar planning depths to fine-tuning, though less robustly. And although we cannot fine-tune GPT-5.4, it fails to exceed seven latent planning steps in the few-shot setting. Together, these results suggest the limitation is unlikely to vanish abruptly in the most capable current models.

Furthermore, while we use large data volumes to detect the discovery bottleneck in open-source models, large-scale fine-tuning of closed-source models is restricted by API costs. Although we cannot rule out that more training data might push the discovery ceiling further, the abrupt nature of the failure (models transition sharply from perfect performance to random guessing as depth increases by one step) and the failure of data scaling to overcome limits in open-source LLMs suggest the bottleneck is unlikely to yield to data scaling alone. An important direction is to systematically map data scaling laws across different depths of implicit search under an unconstrained computation budget.

Finally, our findings are bounded by the strict requirement of fully implicit planning. We show a discovery boundary when models must complete all lookahead computations implicitly, supervised solely by a single cross-entropy loss from node prediction. A natural extension is to explore how supervision signals of varying granularities dynamically alter the depth ceiling of implicit reasoning.

Our findings suggest that current LLMs cannot discover deep latent planning strategies `between tokens` on their own, and that massive scaling alone does little to improve this ability. If this limitation extends beyond planning, models may have no choice but to externalize complex reasoning. Recent work suggests that when chain-of-thought is necessary for task performance, as in our star-graph task, it tends to be faithful to the model’s actual reasoning process \citep{emmons2025chainthoughtnecessarylanguage}. Characterizing the limits of latent reasoning and strategy discovery across domains is a crucial direction for future work that can lend further credence to chain-of-thought monitoring as a viable oversight strategy.

\section*{Acknowledgements}
We thank Boyu Zhu for generously providing computational resources that made this work possible. Further, we thank the Lingo lab at MIT for feedback on an earlier draft that substantially improved this manuscript.

\clearpage

\bibliography{colm2026_conference}
\bibliographystyle{colm2026_conference}
\clearpage
\appendix

\section{Reproducibility Statement}
We provide detailed descriptions of the dataset generation procedure in Appendix~\ref{appendix:dataset_construction}. Complete training hyperparameters for all model architectures are reported in Appendix~\ref{appendix:train}. Dataset statistics for each experimental configuration are summarized in Table~\ref{tab:dataset_stats}. Prompting templates are shown in Appendix~\ref{app:prompts}. All data and scripts will be released publicly
upon acceptance to facilitate reproducibility.

\section{Chain-of-Thought Monitoring}
\label{app:cot_monitoring}
Ensuring the safe deployment of increasingly capable AI systems requires overcoming the inherent opacity of these systems. Given that LLMs articulate intermediate computations in natural language, chain-of-thought (CoT) monitoring provides a critical window into the internal processes of these models \citep{korbak2025chain}. By auditing these explicit reasoning rationales, CoT monitors can preemptively intercept malicious actions \citep{pmlr-v235-greenblatt24a, baker2025monitoring, chennabasappa2025llamafirewall}, identify emerging signs of deceptive alignment \citep{greenblatt2024alignment}, and uncover blind spots in evaluation protocols \citep{meng2025introducing}. However, the viability of CoT monitoring may be inherently fragile, as the reliability of this safety mechanism hinges on a critical distinction: whether a model genuinely requires the externalization of reasoning to succeed on complex tasks, or merely exhibits a propensity to do so, which could be easily overridden by training incentives to hide or shortcut reasoning steps \citep{korbak2025chain}. Our work confirms that externalizing reasoning is indeed a fundamental necessity. By demonstrating a persistent ``depth ceiling'' in latent strategy discovery, we provide empirical evidence that current LLMs fundamentally struggle with complex, multi-step planning entirely within latent states. This limitation dictates that models must rely on externalized CoT for deep reasoning tasks, thereby providing a crucial safety margin for continued CoT monitoring.

\section{Next-Token Prediction}
\label{appendix:ntp}
Given a graph configuration $\mathcal{G}_{(k,m)}$, we construct a synthetic dataset:
\[
\mathcal{D}_{(k,m)} = \{(\mathbf{x}^{(i)}, \mathbf{y}^{(i)})\}_{i=1}^N,
\]
where each input sequence $\mathbf{x}^{(i)}$ encodes $[G_{(k,m)}, v_{\mathrm{source}}, v_{\mathrm{target}}]$ and the target is $\mathbf{y}^{(i)} = [v_{\mathrm{ground}}, \texttt{<eos>}]$ (see Appendices~\ref{appendix:dataset_construction} and \ref{appendix:dataset_stats} for implementation details and statistics). For simplicity, we omit the $\texttt{<eos>}$ token in subsequent sections. The model $\pi_{\theta}$ is trained via standard next-token prediction, minimizing the cross-entropy loss:
\[
\mathcal{L}(\theta)
= \mathbb{E}_{(\mathbf{x}, \mathbf{y}) \sim \mathcal{D}_{(k,m)}}
\left[ - \log \pi_\theta(\mathbf{y} \mid \mathbf{x}) \right].
\]

We train with next-token prediction because reinforcement learning settings typically permit reasoning to be externalized through intermediate tokens, making latent planning difficult to enforce.

\section{Detailed Metrics}
\label{appendix:metric}
In this section, we define metrics to quantify (i) model performance under each graph configuration $\mathcal{G}_{(k,m)}$ and (ii) the effective latent planning depth.

\paragraph{Accuracy} Given a task $\mathcal{G}_{(k,m)}$ and the test set $\hat{\mathcal{D}}_{(k,m)}$, we define \textit{accuracy} as:
\[
\mathrm{Acc}(\pi_\theta, k,m)
= \mathbb{E}_{(\mathbf{x},\mathbf{y}) \sim \hat{\mathcal{D}}_{(k,m)}}
\left[ \mathbb{I}\!\left(v_{\mathrm{pred}} = v_{\mathrm{ground}}\right) \right],
\]
where $v_{\mathrm{pred}} = \arg\max_{v} \pi_\theta(v \mid \mathbf{x})$ under greedy decoding.

For a random guessing baseline that uniformly selects among neighbors of the source node, we have
\[
\mathrm{Acc}(\pi_\mathrm{random}, k, m) = \frac{1}{k}.
\]

\paragraph{Empirical Skill} 
Accuracy alone does not provide a fair comparison across tasks with different branch factors $k$, as the random guessing baseline inherently varies. For instance, an accuracy of $0.7$ represents a marginal improvement over chance when $k=2$, yet indicates substantial predictive capability when $k=10$. To account for this, we define \textit{empirical skill} as the degree to which accuracy exceeds the random baseline:
\begin{equation*}
\text{Skill}(\pi_{\theta}, k, m)=
\dfrac{\mathrm{Acc}(\pi_{\theta}, k, m)-\frac{1}{k}}{1-\frac{1}{k}}.
\end{equation*}
By construction, a skill of $1$ indicates perfect performance and $0$ corresponds to random guessing. 
Note that negative values are possible, with a $k$-dependent lower bound of $\frac{-1}{k-1}$. Since our primary interest is in measuring performance above the random baseline, we intentionally leave the negative range unnormalized and treat all sub-baseline performance uniformly. 

\paragraph{Latent Planning Capability} 
In addition to empirical skill, which measures the exact proficiency of a model, we define the \emph{latent planning capacity} (LPC) at depth $m$ to capture whether a model exhibits any evidence of latent planning.
Formally, we consider a set of branch factors:
\[
\mathcal{K} = \{2, 3, 4, 5, 10\},
\]
where each $k \in \mathcal{K}$ specifies the number of branches in graphs $\mathcal{G}_{(k,m)}$.
A model $\pi_\theta$ is said to possess latent planning capability at depth $m$ if
\[
\mathrm{LPC}(\pi_\theta, m)
=
\begin{cases}
1, &  \max_{k \in \mathcal{K}} \left[ \mathrm{Skill}(\pi_\theta, k,m) - \tau_{\mathrm{crit}}(k, \hat{N}, \alpha = 10^{-5}) \right] \ge 0, \\
0, & \text{otherwise},
\end{cases}
\]
where $\tau_{\mathrm{crit}}(k, \hat{N}, \alpha)$ denotes the \textit{critical skill threshold}, the minimum empirical skill required to reject the null hypothesis of random guessing, given $\hat{N}$ test samples with $k$ branches at significance level $\alpha = 10^{-5}$ (see Table~\ref{tab:acc-skill-grid} for detailed values). Intuitively, $\mathrm{LPC}(\pi_\theta, m) = 1$ as long as the model shows any statistically significant evidence of planning at depth $m$, even if the improvement over random guessing is small.

\section{Strategy Analysis and Attention Probing}

\subsection{Strategy Analysis}
\label{appendix:strategy_analysis}
We identify two principal classes of strategies for solving the star graph task, distinguished by the direction of information propagation.

\paragraph{Forward Strategy}
A forward strategy propagates information from $v_{\mathrm{source}}$ toward $v_{\mathrm{target}}$. The most natural variant is a parallel breadth-first approach, in which the model simultaneously tracks all $k$ branches, advancing one depth level per computational step and checking at each level whether any branch reaches $v_{\mathrm{target}}$. This requires $m$ sequential steps but demands parallel processing across all branches at each step. A sequential depth-first variant, which fully traverses one branch before returning to try the next, is less plausible: it requires up to $O(k \cdot m)$ steps in the worst case, as the model must explore and abandon incorrect branches one at a time. The sole exception arises when $k=2$, where the model can randomly select one branch and, upon failing to reach $v_{\mathrm{target}}$, identify the correct branch through elimination.

\paragraph{Backtracking Strategy}
A backtracking strategy propagates information in the reverse direction, tracing the path from $v_{\mathrm{target}}$ back to $v_{\mathrm{source}}$. Since $v_{\mathrm{target}}$ is a leaf node with a unique path to the root, this strategy requires exactly $m$ sequential steps and does not require parallel tracking of multiple branches or trial-and-error traversal. We consider this the most efficient strategy, as it avoids both the parallel processing demands of forward breadth-first search and the worst-case overhead of forward depth-first search.

\subsection{Strategy Probing via Attention in Transformers}
\label{appendix:attention_probing}
Since the transformer is initialized without any pre-existing knowledge, its attention patterns reflect the unconfounded strategies acquired during training. Within this controlled setting, we treat the attention weights as a probe to determine whether the model selectively attends to relevant graph components during inference.

\paragraph{Input for Transformers}
Specifically, we encode the input $[G_{(k,m)}, v_{\mathrm{source}}, v_{\mathrm{target}}]$ as a token sequence
\[
\mathbf{x} = [u_1, v_1, s, \ldots, u_M, v_M, g, v_{\mathrm{source}}, v_{\mathrm{target}}, g],
\]
where $(u_i, v_i)$ are node tokens forming edges, $s$ is a separator token between edges, and $g$ is the token that delineates the graph representation from $(v_{\mathrm{source}}, v_{\mathrm{target}})$ and marks the end of the input for prediction (see Figure \ref{fig:lpc_bar} (right) for an example).

\paragraph{Final Token Attention} 
In an autoregressive model, given $T = |\mathbf{x}|$, the hidden representation of the last input token is used to predict the next token. Therefore, we simply focus on the attention
computed by this token.

For each transformer layer $\ell$, we extract the attention distribution
\[
\mathbf{a}^{(\ell)} = \mathrm{softmax}\!\left(
\frac{\mathbf{q}^{(\ell)} \mathbf{K}^{(\ell)\top}}{\sqrt{d}}
\right),
\]
where $\mathbf{q}^{(\ell)}$ is the query vector of the final token at layer $\ell$, $\mathbf{K}^{(\ell)}$ denotes the key vectors of all input tokens, and $d$ is the head dimension. Each token thus receives a scalar attention weight indicating its influence on the predicted next-hop node.

\paragraph{Token Partition}
Let $x_t$ denote the token at position $t$ in the input sequence $\mathbf{x}$. We partition the input tokens into three disjoint subsets:
$\mathcal{T}_{\mathrm{problem}}=\{\,t\mid x_t\in\{v_{\mathrm{source}},v_{\mathrm{target}}\}\,\}$,
$\mathcal{T}_{\mathrm{edge}}=\{\,t\mid x_t\in\{u_i,v_i\}_{i=1}^M\,\}$,
and
$\mathcal{T}_{\mathrm{special}}=\{\,t\mid x_t\in\{s,g\}\,\}$.

\paragraph{Backtracking Ratio}
We interpret attention on edge tokens as a signal of which edges are actively considered by the model.
If the model follows a parallel strategy such as breadth-first search (BFS) or blind search, attention over edges is likely to be approximately uniform.
In contrast, a backtracking strategy that traverses the branch in reverse order likely concentrates attention on edges along the path between $v_{\mathrm{target}}$ and $v_{\mathrm{source}}$.

In this case, we categorize edge tokens into
$\mathcal{T}_{\mathrm{on}}$ and
$\mathcal{T}_{\mathrm{off}}$,
where $\mathcal{T}_{\mathrm{on}}$ consists of node tokens belonging to edges on the path from $v_{\mathrm{target}}$ to $v_{\mathrm{source}}$, and
$\mathcal{T}_{\mathrm{off}} = \mathcal{T}_{\mathrm{edge}} \setminus \mathcal{T}_{\mathrm{on}}$.

Subsequently, we define the \emph{backtracking ratio} (BR) as the fraction of edge-token attention on $\mathcal{T}_{\mathrm{on}}$:
\[
\mathrm{BR}
=
\frac{
\sum_{\ell,\; t \in \mathcal{T}_{\mathrm{on}}} a^{(\ell)}_t
}{
\sum_{\ell,\; t \in \mathcal{T}_{\mathrm{on}}} a^{(\ell)}_t
+
\sum_{\ell,\; t \in \mathcal{T}_{\mathrm{off}}} a^{(\ell)}_t
},
\]
where $a^{(\ell)}_t$ denotes the $t$-th entry of $\mathbf{a}^{(\ell)}$. If the model adopts a backtracking strategy, the $\mathrm{BR}$ will be considerably above $\frac{1}{k}$.

\section{Dataset}
\subsection{Dataset Construction}
\label{appendix:dataset_construction}
Given a specific configuration $\mathcal{G}_{(k,m)}$, we first generate an undirected star graph comprising a central node and $k$ outward radiating branches, each of length $m$. To prevent the model from exploiting numerical patterns, we randomly permute all node identifiers. This ensures that the model must infer the topological structure solely from relational connectivity. We then designate the central node as the source and the terminal node of a randomly selected branch as the target.

Next, we serialize the graph into a list of independent edges, where each edge is represented by a pair of integer node labels. We randomly shuffle the order of these edges. This permutation forces the model to comprehend the true underlying connectivity rather than relying on the sequential positioning of edges within the input.

Finally, we implement a strict filtering mechanism to partition the data into mutually disjoint training, validation, and test sets. We first generate the validation and test samples, recording their encoded representation. During the generation of the training set, we compare every new sample against these recorded configurations. We immediately discard and regenerate any training sample that matches an existing validation or test instance, thereby guaranteeing absolute mutual exclusivity and preventing data leakage across the evaluation splits.

\subsection{Dataset Statistics}
\label{appendix:dataset_stats}
We scale the dataset sizes according to the computational cost of each model architecture. For the from-scratch Transformer, we use 100000 samples for training, 2048 for validation, and 2048 for testing. When fine-tuning open-weight LLMs, we adjust the training size to balance sample efficiency and computational overhead. Specifically, we utilize 10000 training samples for Qwen-2.5 7B and Qwen-3 8B, and 2000 training samples for the larger 32B variants. We evaluate all Qwen models on a fixed test set of 2048 samples. For the proprietary GPT-4o model, we restrict the training to 100 samples (this is also the recommended size from OpenAI). Due to the computational cost considerations, we limit the test set size for both GPT-4o and GPT-5.4 to 100 samples either. Table \ref{tab:dataset_stats} summarizes the exact dataset splits used across all experiments.

\begin{table}
\centering
\caption{Dataset statistics across models.}
\label{tab:dataset_stats}
\begin{tabular}{lccc}
\toprule
Model & Train Size & Validation Size & Test Size \\
\midrule
Transformer & 100000 & 2048 & 2048 \\
Qwen 2.5 & & & \\
\phantom{xxx}- 7B & 10000 & N/A & 2048 \\
\phantom{xxx}- 32B & 2000 & N/A & 2048 \\
Qwen 3 & & & \\
\phantom{xxx}- 8B & 10000 & N/A & 2048 \\
\phantom{xxx}- 32B & 2000 & N/A & 2048 \\
GPT-4o & 100 & N/A & 100 \\
GPT-5.4 & N/A & N/A & 100 \\
\bottomrule
\end{tabular}
\end{table}

\section{Training Details}
\label{appendix:train}
\subsection{Models}
\paragraph{From-Scratch Transformer} To isolate strategy discovery from pre-training effects, we train an autoregressive transformer following the standard GPT-2 architecture \citep{radford2019language} from scratch, using GELU \citep{hendrycks2016gaussian} as the activation function. Specifically, our default configuration consists of 8 layers, 1 attention head, and a hidden dimension of 128. Furthermore, we conducted an ablation study on these architectural hyperparameters. As shown in Table~\ref{tab:ablation}, scaling up the model by increasing the number of layers, attention heads, or hidden dimensions does not yield any improvement in the strategy discovery depth ceiling, confirming that the failure to discover deeper strategies is not simply due to a lack of basic representational capacity.

\paragraph{Open-Source LLMs} For all open-source models, we exclusively utilize the base pre-trained checkpoints. Specifically, we use Qwen 2.5-7B (Base), Qwen 2.5-32B (Base), Qwen 3-8B (Base), and Qwen 3-32B (Base).

\paragraph{Closed-Source LLMs} For the closed-source models, we use \texttt{gpt-4o-2024-08-06} for both evaluation and fine-tuning via the OpenAI API, and \texttt{gpt-5.4-2026-03-05} for evaluation only, as fine-tuning is not publicly available for this model.

\subsection{Hyperparameters}
We summarize the comprehensive hyperparameters utilized for training across all experiments in Table~\ref{tab:hyperparameters}. For the training of the from-scratch transformer, we adopt the Adam optimizer \citep{kingma2014adam} without weight decay. To further enhance training stability and generalization, we apply a dropout rate of $0.1$. For the fine-tuning of open-source LLMs, we utilize the default AdamW optimizer \citep{loshchilov2017decoupled}. Regarding the closed-source model, GPT-4o, the learning rate and the choice of optimizer are automatically determined by the API due to the black-box nature of the system.

\begin{table}[t]
\centering
\caption{Hyper-parameters for training models on star graph tasks. Note that the reported batch size represents the per-GPU batch size.}
\vspace{2mm}
\renewcommand{\arraystretch}{1.2}
\resizebox{\linewidth}{!}{
\begin{tabular}{l ccccc}
\toprule
\multirow{2}{*}{\textbf{Hyper-parameter}}
& \multicolumn{2}{c}{\textbf{From-Scratch Transformer}}
& \multicolumn{2}{c}{\textbf{Open-Source LLMs}}
& \textbf{Closed-Source} \\
\cmidrule(lr){2-3} \cmidrule(lr){4-5} \cmidrule(lr){6-6}
& Latent / CoT & ICoT & 7B / 8B & 32B & GPT-4o \\
\midrule
Epochs        & 500                & 1000               & 1                  & 1                  & 10 \\
Learning Rate & $1 \times 10^{-4}$ & $1 \times 10^{-4}$ & $3 \times 10^{-6}$ & $3 \times 10^{-6}$ & Auto \\
Batch Size    & 1024               & 1024               & 4                  & 1                  & 1 \\
Dropout       & 0.1                & 0.1                & N/A                & N/A                 & - \\
GPUs          & 1                  & 1                  & 8                  & 8                  & - \\
\bottomrule
\end{tabular}
}
\label{tab:hyperparameters}
\end{table}

\begin{table*}[t!]
\centering
\small
\renewcommand{\arraystretch}{1.15}
\setlength{\tabcolsep}{5pt}
\caption{
In-distribution skill of trained models on star graph tasks across planning depths $m \in \{3,4,5,6\}$ and branch factors $k \in \{2,3,4,5,10\}$. 
Each model is trained and evaluated on the same graph configuration $\mathcal{G}_{(k,m)}$. 
\textsuperscript{†} denotes trained models.
\win{Blue-highlighted bold text} indicates configurations where the model successfully rejects random guessing, while \fail{gray text} indicates failure.
}
\label{tab:acc_trained}
\begin{adjustbox}{max width=\textwidth}
\resizebox{\textwidth}{!}{%
\begin{tabular}{l ccccc|ccccc|ccccc|ccccc}
\toprule
\multirow{2}{*}{Model}
& \multicolumn{5}{c}{$m=3$}
& \multicolumn{5}{c}{$m=4$}
& \multicolumn{5}{c}{$m=5$}
& \multicolumn{5}{c}{$m=6$} \\
\cmidrule(lr){2-6}\cmidrule(lr){7-11}\cmidrule(lr){12-16}\cmidrule(lr){17-21}
& 2 & 3 & 4 & 5 & 10
& 2 & 3 & 4 & 5 & 10
& 2 & 3 & 4 & 5 & 10
& 2 & 3 & 4 & 5 & 10 \\
\midrule
\multicolumn{21}{c}{From-Scratch Model} \\ 
\midrule \midrule
Transformer\textsuperscript{†} 
& \win{1.00} & \win{1.00} & \win{1.00} & \win{1.00} & \fail{-0.04}
& \fail{0.02} & \fail{0.03} & \fail{0.00} & \fail{0.00} & \fail{-0.07}
& \fail{--} & \fail{--} & \fail{--} & \fail{--} & \fail{--}
& \fail{--} & \fail{--} & \fail{--} & \fail{--} & \fail{--} \\
\midrule
\multicolumn{21}{c}{Open-Source Model} \\ 
\midrule \midrule
Qwen 2.5 & & & & & & & & & & & & & & & & & & & & \\
\phantom{xxx}- 7B\textsuperscript{†} 
& \win{1.00} & \win{1.00} & \win{1.00} & \win{1.00} & \win{1.00}
& \fail{-0.04} & \fail{-0.01} & \fail{0.00} & \fail{-0.01} & \fail{-0.01}
& \fail{--} & \fail{--} & \fail{--} & \fail{--} & \fail{--}
& \fail{--} & \fail{--} & \fail{--} & \fail{--} & \fail{--} \\
\phantom{xxx}- 32B\textsuperscript{†} 
& \win{1.00} & \win{1.00} & \win{1.00} & \win{1.00} & \win{0.97}
& \win{0.98} & \win{1.00} & \win{1.00} & \win{0.99} & \win{1.00}
& \fail{0.00} & \fail{-0.01} & \fail{-0.01} & \fail{0.00} & \fail{0.01}
& \fail{--} & \fail{--} & \fail{--} & \fail{--} & \fail{--} \\
Qwen 3 & & & & & & & & & & & & & & & & & & & & \\
\phantom{xxx}- 8B\textsuperscript{†} 
& \win{1.00} & \win{1.00} & \win{1.00} & \win{1.00} & \win{1.00}
& \fail{0.00} & \win{1.00} & \win{1.00} & \win{1.00} & \win{0.99}
& \fail{0.00} & \fail{0.04} & \fail{-0.01} & \fail{0.00} & \fail{0.01}
& \fail{--} & \fail{--} & \fail{--} & \fail{--} & \fail{--} \\
\phantom{xxx}- 32B\textsuperscript{†} 
& \win{1.00} & \win{1.00} & \win{0.97} & \win{1.00} & \win{1.00}
& \win{1.00} & \win{0.99} & \win{0.96} & \win{0.99} & \win{1.00}
& \fail{0.00} & \fail{-0.01} & \win{0.99} & \win{0.99} & \fail{-0.01}
& \fail{-0.02} & \fail{-0.02} & \fail{0.01} & \fail{0.00} & \fail{0.00} \\
\midrule
\multicolumn{21}{c}{Closed-Source Model} \\ 
\midrule \midrule
GPT-4o\textsuperscript{†} 
& \win{1.00} & \win{1.00} & \win{1.00} & \win{1.00} & \win{1.00}
& \win{1.00} & \win{1.00} & \win{1.00} & \win{0.99} & \win{0.98}
& \win{1.00} & \win{0.96} & \fail{-0.05} & \win{1.00} & \win{0.98}
& \fail{0.08} & \fail{-0.02} & \fail{-0.11} & \fail{0.05} & \fail{-0.06} \\
\bottomrule
\end{tabular}%
}
\end{adjustbox}
\end{table*}

\section{Transformer Results}

\subsection{Training Dynamics of Transformer}

\begin{figure*}[t]
  \centering
  \includegraphics[width=1.0\textwidth]{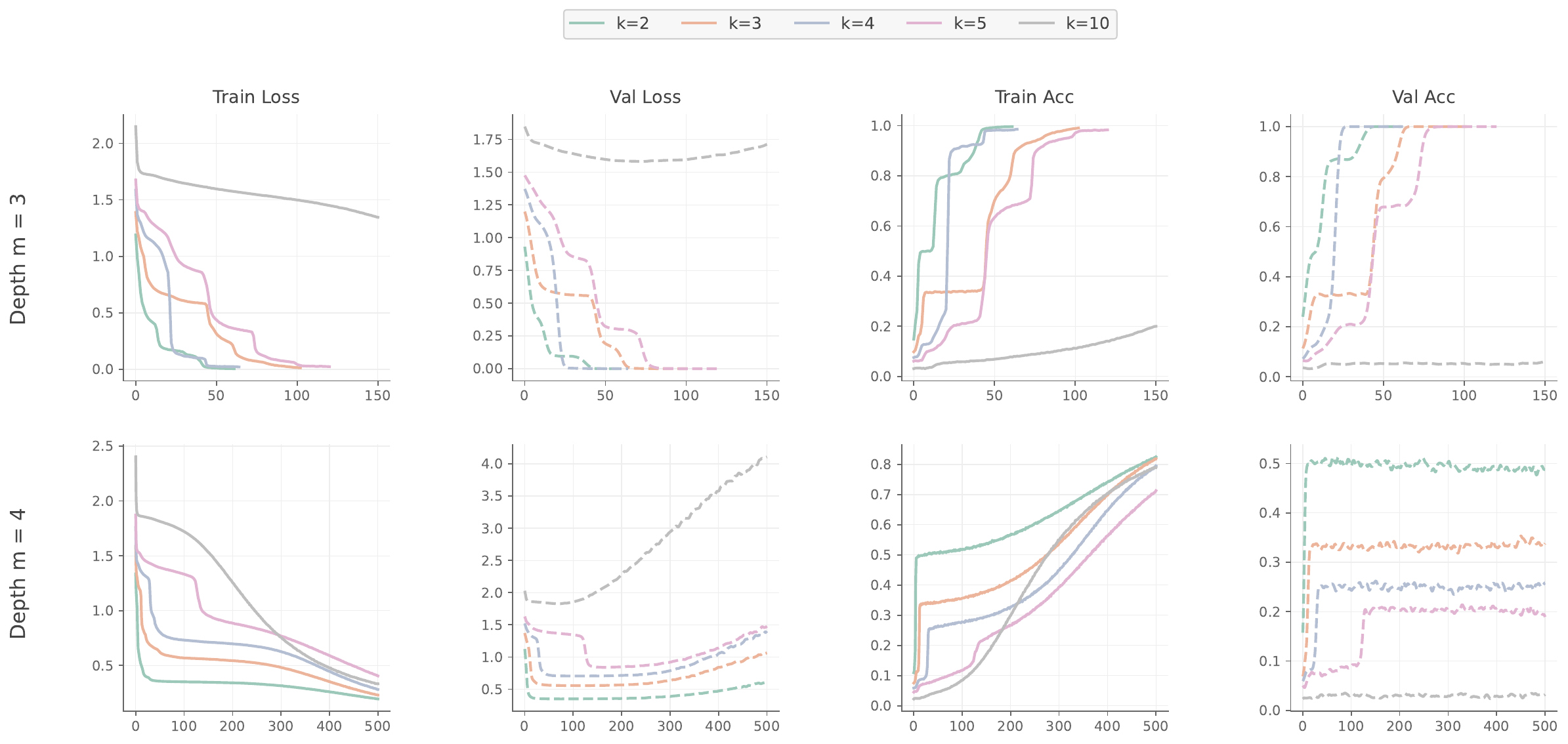}
    \caption{\textbf{Training dynamics of the from-scratch transformer across planning depths $m \in \{3, 4\}$ and branch factors $k \in \mathcal{K}$.} The model exhibits a two-stage learning process: validation accuracy first rises to the random baseline ($\frac{1}{k}$) as the model learns to predict valid neighbors, then either jumps to near-perfect accuracy (successful strategy discovery) or stagnates while training loss continues to decrease (overfitting without strategy discovery).}
    \label{fig:transfomer_loss}
\end{figure*}

\begin{table*}[t!]
\centering
\small
\renewcommand{\arraystretch}{1.15}
\setlength{\tabcolsep}{5pt}
\caption{
Ablation study on Transformer architecture hyperparameters for star graph tasks across planning depths $m \in \{3,4\}$ and branch factors $k \in \{2,3,4,5,10\}$.
The default configuration is (8 layers, 128 dim, 1 head). Each ablation varies one hyperparameter while keeping the others at default.
\textsuperscript{†} denotes trained models.
\win{Blue-highlighted bold text} indicates configurations where the model successfully rejects random guessing, while \fail{gray text} indicates failure.
}
\label{tab:ablation}
\begin{adjustbox}{max width=\textwidth}
\begin{tabular}{l ccccc|ccccc}
\toprule
\multirow{2}{*}{Model}
& \multicolumn{5}{c}{$m=3$}
& \multicolumn{5}{c}{$m=4$} \\
\cmidrule(lr){2-6}\cmidrule(lr){7-11}
& 2 & 3 & 4 & 5 & 10
& 2 & 3 & 4 & 5 & 10 \\
\midrule
\multicolumn{11}{c}{\textit{Ablation on Layers} (dim=128, heads=1)} \\
\midrule \midrule
Transformer & & & & & & & & & & \\
\phantom{xxx}- 16 layers\textsuperscript{†}
& \win{1.00} & \win{1.00} & \win{1.00} & \win{1.00} & \fail{-0.06}
& \fail{0.00} & \fail{0.00} & \fail{-0.01} & \fail{-0.17} & \fail{-0.08} \\
\phantom{xxx}- 32 layers\textsuperscript{†}
& \win{1.00} & \win{1.00} & \win{1.00} & \win{1.00} & \fail{-0.05}
& \fail{-0.01} & \fail{-0.02} & \fail{-0.02} & \fail{0.00} & \fail{-0.08} \\
\midrule
\multicolumn{11}{c}{\textit{Ablation on Heads} (layers=8, dim=128)} \\
\midrule \midrule
Transformer & & & & & & & & & & \\
\phantom{xxx}- 2 heads\textsuperscript{†}
& \win{1.00} & \win{1.00} & \win{1.00} & \win{1.00} & \fail{-0.07}
& \fail{0.00} & \fail{0.02} & \fail{-0.02} & \fail{0.02} & \fail{-0.08} \\
\phantom{xxx}- 4 heads\textsuperscript{†}
& \win{0.93} & \win{0.98} & \win{1.00} & \win{1.00} & \fail{-0.06}
& \fail{0.03} & \fail{0.02} & \fail{-0.04} & \fail{-0.13} & \fail{-0.08} \\
\phantom{xxx}- 8 heads\textsuperscript{†}
& \win{1.00} & \win{0.75} & \win{1.00} & \fail{-0.01} & \fail{-0.06}
& \fail{-0.01} & \fail{0.00} & \fail{-0.02} & \fail{-0.16} & \fail{-0.08} \\
\midrule
\multicolumn{11}{c}{\textit{Ablation on Dimension} (layers=8, heads=1)} \\
\midrule \midrule
Transformer & & & & & & & & & & \\
\phantom{xxx}- 256 dim\textsuperscript{†}
& \win{1.00} & \win{1.00} & \win{1.00} & \win{1.00} & \fail{-0.03}
& \fail{-0.02} & \fail{0.04} & \fail{-0.02} & \fail{0.00} & \fail{-0.08} \\
\phantom{xxx}- 512 dim\textsuperscript{†}
& \win{1.00} & \win{1.00} & \win{1.00} & \fail{-0.13} & \fail{-0.08}
& \fail{0.04} & \fail{0.01} & \fail{0.00} & \fail{-0.18} & \fail{-0.06} \\
\bottomrule
\end{tabular}
\end{adjustbox}
\end{table*}

\begin{table*}[t!]
\centering
\small
\renewcommand{\arraystretch}{1.15}
\setlength{\tabcolsep}{5pt}
\caption{
Zero-shot and few-shot skill of pre-trained models on star graph tasks across planning depths $m \in \{3,4,5,6\}$ and branch factors $k \in \{2,3,4,5,10\}$. No task-specific training is applied.
\win{Blue-highlighted bold text} indicates configurations where the model successfully rejects random guessing, while \fail{gray text} indicates failure.
}
\label{tab:acc_untrained}
\begin{adjustbox}{max width=\textwidth}
\resizebox{\textwidth}{!}{%
\begin{tabular}{l ccccc|ccccc|ccccc|ccccc}
\toprule
\multirow{2}{*}{Model}
& \multicolumn{5}{c}{$m=3$}
& \multicolumn{5}{c}{$m=4$}
& \multicolumn{5}{c}{$m=5$}
& \multicolumn{5}{c}{$m=6$} \\
\cmidrule(lr){2-6}\cmidrule(lr){7-11}\cmidrule(lr){12-16}\cmidrule(lr){17-21}
& 2 & 3 & 4 & 5 & 10
& 2 & 3 & 4 & 5 & 10
& 2 & 3 & 4 & 5 & 10
& 2 & 3 & 4 & 5 & 10 \\
\midrule
\multicolumn{21}{c}{Zero Shot Performance} \\ 
\midrule \midrule
Qwen 2.5 & & & & & & & & & & & & & & & & & & & & \\
\phantom{xxx}- 32B
& \win{0.30} & \win{0.25} & \win{0.28} & \win{0.22} & \win{0.16}
& \fail{0.00} & \fail{0.04} & \fail{0.05} & \fail{0.01} & \fail{0.01}
& \fail{-0.02} & \fail{-0.02} & \fail{-0.01} & \fail{0.00} & \fail{0.00}
& \fail{-0.02} & \fail{-0.01} & \fail{0.01} & \fail{-0.02} & \fail{0.01} \\
Qwen 3 & & & & & & & & & & & & & & & & & & & & \\
\phantom{xxx}- 8B
& \fail{0.02} & \win{0.16} & \win{0.20} & \win{0.18} & \win{0.14}
& \fail{-0.10} & \fail{-0.03} & \fail{0.00} & \fail{-0.01} & \fail{0.00}
& \fail{-0.18} & \fail{-0.03} & \fail{-0.04} & \fail{-0.02} & \fail{0.00}
& \fail{-0.10} & \fail{-0.05} & \fail{-0.05} & \fail{-0.01} & \fail{-0.01} \\
\phantom{xxx}- 32B
& \win{0.14} & \win{0.23} & \win{0.25} & \win{0.30} & \win{0.24}
& \fail{0.00} & \win{0.07} & \fail{0.03} & \win{0.12} & \win{0.06}
& \fail{-0.02} & \fail{0.02} & \fail{0.05} & \fail{0.01} & \fail{0.00}
& \fail{-0.02} & \fail{0.01} & \fail{-0.03} & \fail{0.01} & \fail{-0.01} \\
GPT-4o
& \win{0.68} & \win{0.62} & \win{0.77} & \win{0.74} & \win{0.57}
& \fail{0.22} & \fail{0.12} & \win{0.35} & \win{0.28} & \fail{0.12}
& \fail{-0.08} & \fail{0.16} & \fail{0.03} & \fail{-0.02} & \fail{-0.03}
& \fail{0.08} & \fail{-0.10} & \fail{-0.01} & \fail{-0.06} & \fail{-0.03} \\
GPT-5.4
& \win{0.90} & \win{0.97} & \win{0.93} & \win{0.98} & \win{1.00}
& \win{0.88} & \win{0.78} & \win{0.87} & \win{0.88} & \win{0.88}
& \fail{0.28} & \win{0.54} & \win{0.51} & \win{0.44} & \win{0.51}
& \fail{0.10} & \fail{0.20} & \fail{0.23} & \fail{0.23} & \fail{0.17} \\
\midrule
\multicolumn{21}{c}{Few Shot Performance} \\ 
\midrule \midrule
Qwen 2.5 & & & & & & & & & & & & & & & & & & & & \\
\phantom{xxx}- 32B
& \win{0.34} & \win{0.54} & \win{0.59} & \win{0.45} & \win{0.42} 
& \fail{0.08} & \win{0.18} & \win{0.20} & \win{0.14} & \win{0.06} 
& \fail{0.00} & \fail{0.02} & \fail{0.04} & \fail{0.02} & \fail{0.00}
& \fail{0.04} & \fail{0.01} & \fail{0.00} & \fail{0.02} & \fail{0.01} \\
Qwen 3 & & & & & & & & & & & & & & & & & & & & \\
\phantom{xxx}- 8B
& \fail{0.04} & \win{0.34} & \win{0.35} & \win{0.28} & \win{0.10}
& \fail{-0.16} & \fail{-0.06} & \fail{0.00} & \fail{0.00} & \fail{-0.03}
& \fail{-0.16} & \fail{-0.05} & \fail{-0.04} & \fail{-0.05} & \fail{-0.02}
& \fail{-0.08} & \fail{-0.08} & \fail{-0.04} & \fail{-0.05} & \fail{0.01} \\
\phantom{xxx}-32B
& \win{0.38} & \win{0.58} & \win{0.57} & \win{0.59} & \win{0.40} 
& \win{0.14} & \win{0.20} & \win{0.17} & \win{0.16} & \win{0.09} 
& \fail{-0.02} & \fail{0.01} & \fail{0.01} & \fail{-0.01} & \fail{-0.01} 
& \fail{-0.04} & \fail{-0.05} & \fail{-0.05} & \fail{-0.04} & \fail{-0.02} \\
GPT-4o
& \win{0.98} & \win{0.94} & \win{0.99} & \win{0.93} & \win{0.80}
& \win{0.60} & \win{0.49} & \win{0.56} & \win{0.46} & \fail{0.07}
& \fail{0.24} & \fail{0.16} & \fail{0.07} & \fail{-0.07} & \fail{-0.03}
& \fail{0.10} & \fail{-0.09} & \fail{-0.05} & \fail{-0.04} & \fail{-0.06}\\
GPT-5.4
& \win{0.98} & \win{1.00} & \win{1.00} & \win{0.99} & \win{1.00}
& \win{0.98} & \win{1.00} & \win{1.00} & \win{0.99} & \win{1.00}
& \win{0.80} & \win{0.70} & \win{0.92} & \win{0.89} & \win{0.81}
& \win{0.66} & \win{0.76} & \win{0.75} & \win{0.61} & \win{0.36} \\
\bottomrule
\end{tabular}%
}
\end{adjustbox}
\end{table*}

\begin{table}[t!]
\centering
\small
\setlength{\tabcolsep}{6pt}
\caption{
To further probe the latent planning capability (LPC) of GPT-5.4, we additionally evaluate it on more challenging star graph tasks with planning depths $m \in \{7,8\}$ and branch factors $k \in \{2,3,4,5,10\}$. We report skill scores for zero-shot and few-shot settings.
\win{Blue-highlighted bold text} indicates configurations where the model successfully rejects random guessing, while \fail{gray text} indicates failure.
}
\label{tab:gpt54_skill_m78}
\begin{tabular}{l ccccc|ccccc}
\toprule
\multirow{2}{*}{Model}
& \multicolumn{5}{c}{$m=7$}
& \multicolumn{5}{c}{$m=8$} \\
\cmidrule(lr){2-6}\cmidrule(lr){7-11}
& 2 & 3 & 4 & 5 & 10
& 2 & 3 & 4 & 5 & 10 \\
\midrule
GPT-5.4 & & & & & & & & & & \\
\phantom{xxx}- Zero Shot
& \fail{0.22} & \fail{0.07} & \fail{0.17} & \fail{-0.03} & \fail{0.02}
& \fail{-0.08} & \fail{0.07} & \fail{-0.03} & \fail{-0.03} & \fail{-0.01} \\
\phantom{xxx}- Few Shot
& \win{0.50} & \fail{0.25} & \fail{0.27} & \fail{0.23} & \fail{0.09}
& \fail{0.00} & \fail{0.12} & \fail{0.17} & \fail{0.01} & \fail{0.06} \\
\bottomrule
\end{tabular}
\end{table}

To understand the training behavior of the from-scratch transformer, we record the loss and accuracy on both the training set and the validation set at each training update. Figure~\ref{fig:transfomer_loss} presents these curves for planning depths $m \in \{3, 4\}$ across all branch factors $k \in \mathcal{K}$.

Across configurations, the model typically exhibits a two-stage learning process. In the first stage, the model learns a simple local heuristic: predicting any valid neighbor of $v_{\mathrm{source}}$. During this stage, the training loss decreases steadily, and the validation accuracy rises to approximately $\frac{1}{k}$, matching the random baseline. The number of training steps required to complete this stage generally increases with the branch factor $k$, reflecting the greater difficulty of learning the local connectivity structure in graphs with more branches. When $k = 10$, the model fails to complete even this first stage, indicating a fundamental capacity limitation in processing complex local connectivity rather than a failure in strategy discovery.

In the second stage, the model attempts to learn the full multi-step planning strategy. For configurations where the model succeeds, the validation accuracy rapidly rises from the random baseline to near-perfect levels, accompanied by a sharp drop in validation loss. This transition is notably abrupt, suggesting that the model discovers the complete planning strategy in a discrete step rather than gradually improving.

For configurations where the model fails to discover the planning strategy, a distinctive pattern emerges: the training loss continues to decrease while the validation loss begins to increase, and the validation accuracy remains at the random baseline. This divergence indicates that the model resorts to memorizing the specific graph instances in the training set rather than learning a generalizable strategy. Since each star graph is constructed with randomly permuted node identifiers, the memorized mappings do not transfer to unseen graphs, leaving validation performance at chance level.

\subsection{Ablation Study on Transformer Hyperparameters}
\label{appendix:hyperparameters}
To verify that the discovery ceiling observed at $m=4$ is not simply due to insufficient model capacity, we conduct an ablation study by independently scaling each architectural hyperparameter of the default transformer ($8$ layers, $128$ hidden dimension, $1$ attention head). Specifically, we increase the number of layers $\{16, 32\}$, attention heads $\{2, 4, 8\}$, and hidden dimension $\{256, 512\}$, while keeping all other hyperparameters at their default values.

As shown in Table~\ref{tab:ablation}, none of these modifications overcome the discovery ceiling. All variants achieve near-perfect skill at $m=3$ for $k < 5$, matching the default configuration. However, at $m=4$, every variant produces skill scores near or below zero across all branch factors, indicating a complete failure to discover the planning strategy at this depth. Notably, increasing the number of attention heads to $8$ even degrades performance at $m=3$, with skill dropping to $0.75$ at $k=3$ and failing entirely at $k=5$. Similarly, increasing the hidden dimension to $512$ leads to failure at $k=5$ for $m=3$. These results suggest that naive architectural scaling does not address the discovery bottleneck, and may even introduce optimization difficulties that impair performance at shallower depths.

\subsection{Attention Visualization}
\label{appendix:attention_visualization}
\begin{figure*}[t]
    \centering

    \begin{minipage}{0.19\textwidth}
        \centering
        \includegraphics[width=\linewidth]{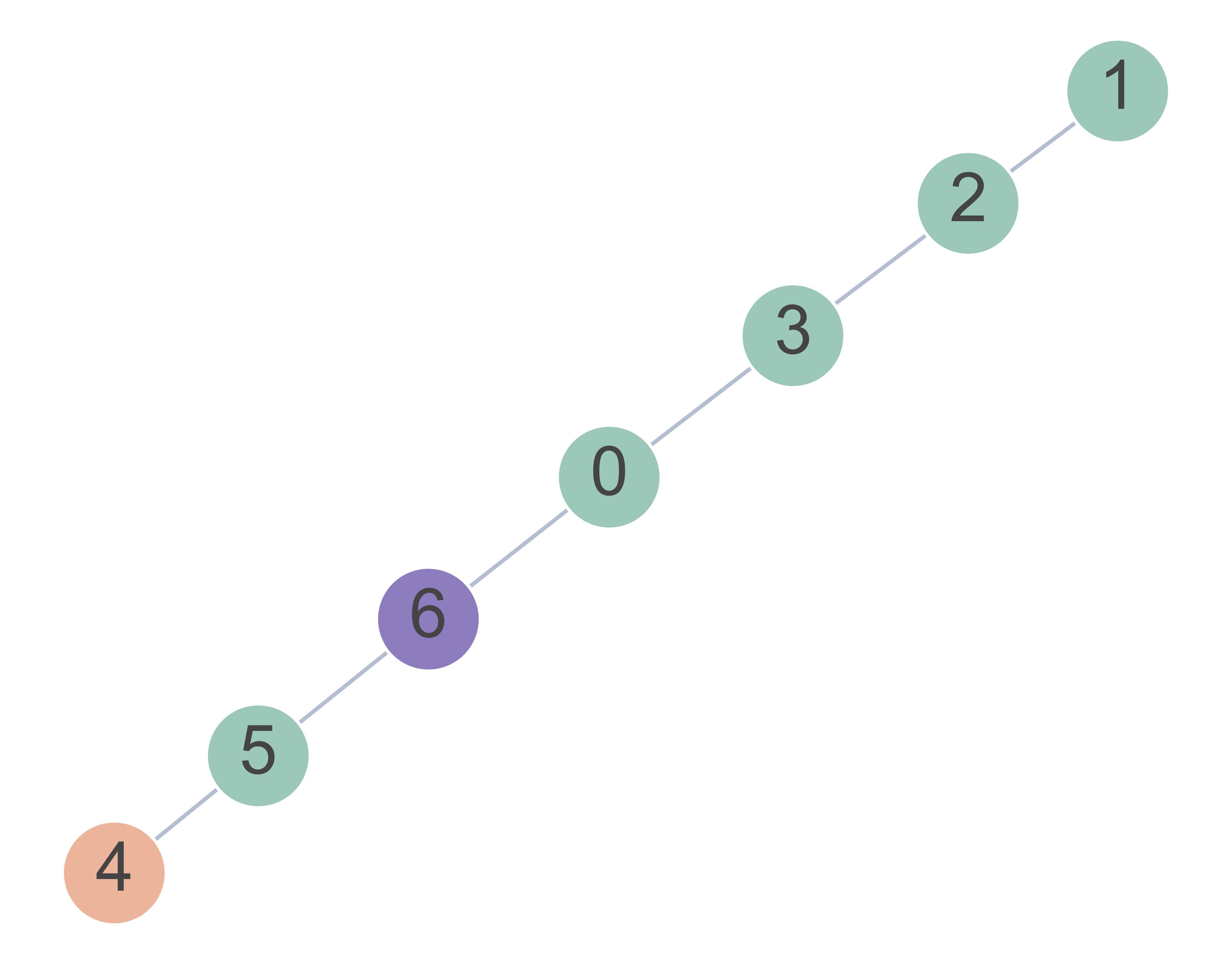}
    \end{minipage}
    \hfill
    \begin{minipage}{0.79\textwidth}
        \centering
        \includegraphics[width=\linewidth]{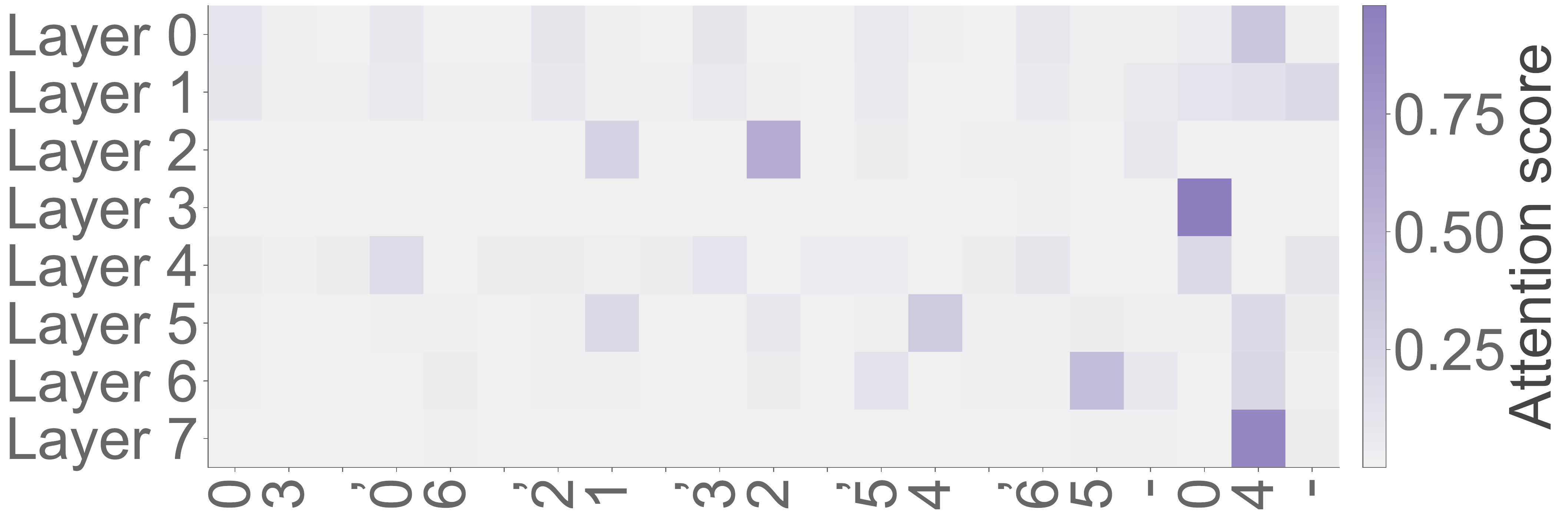}
    \end{minipage}

    \vspace{2em}

    \begin{minipage}{0.19\textwidth}
        \centering
        \includegraphics[width=\linewidth]{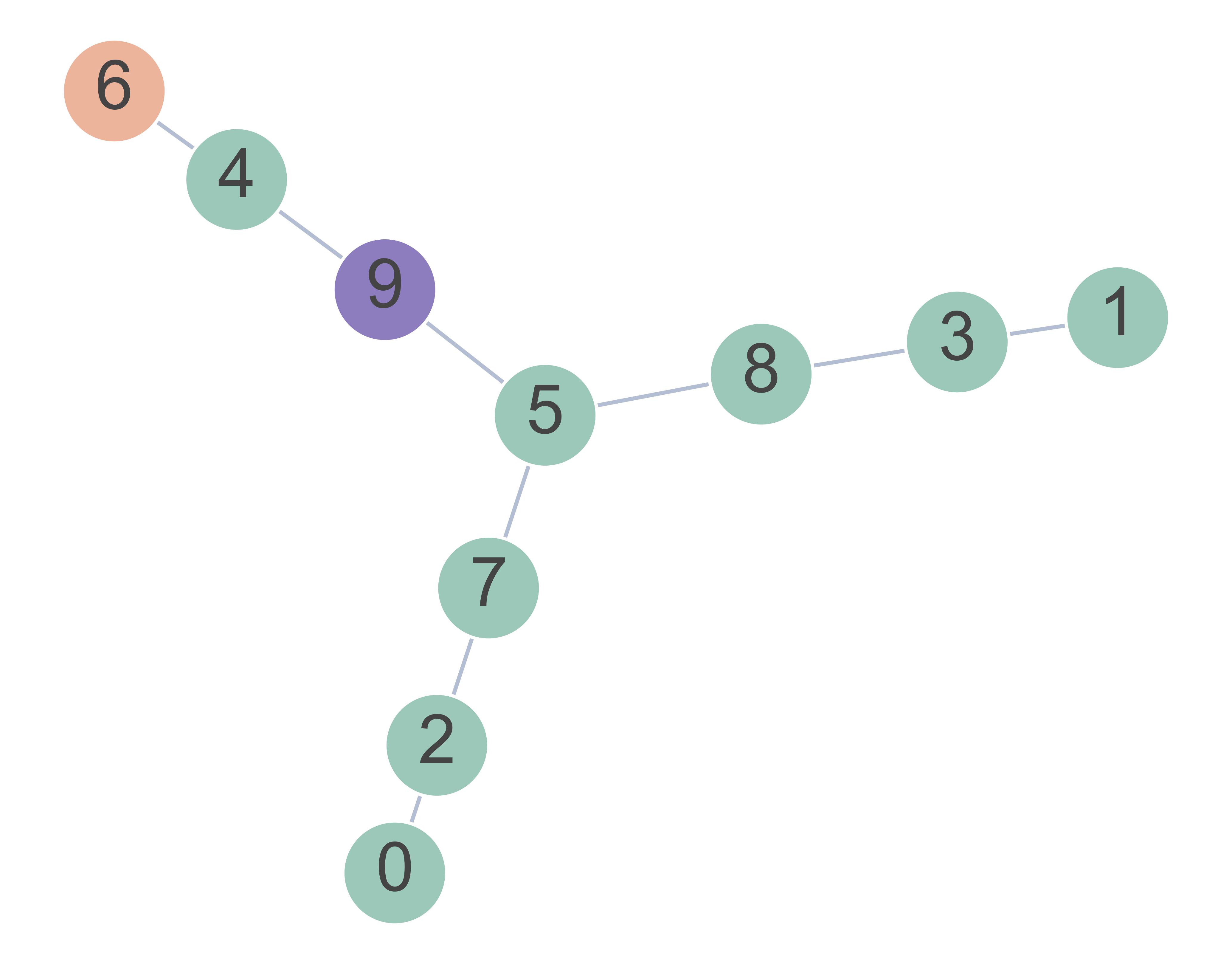}
    \end{minipage}
    \hfill
    \begin{minipage}{0.79\textwidth}
        \centering
        \includegraphics[width=\linewidth]{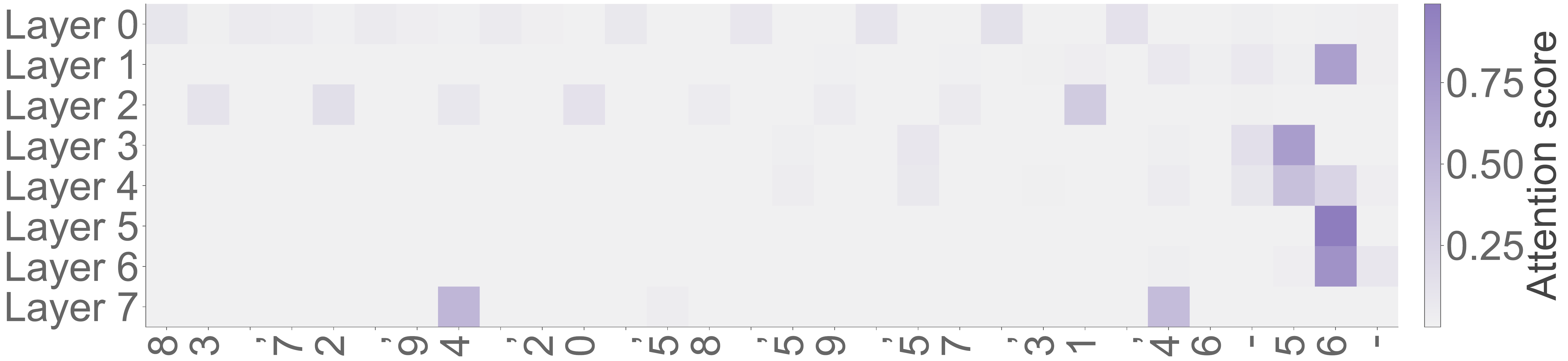}
    \end{minipage}

    \vspace{2em}

    \begin{minipage}{0.19\textwidth}
        \centering
        \includegraphics[width=\linewidth]{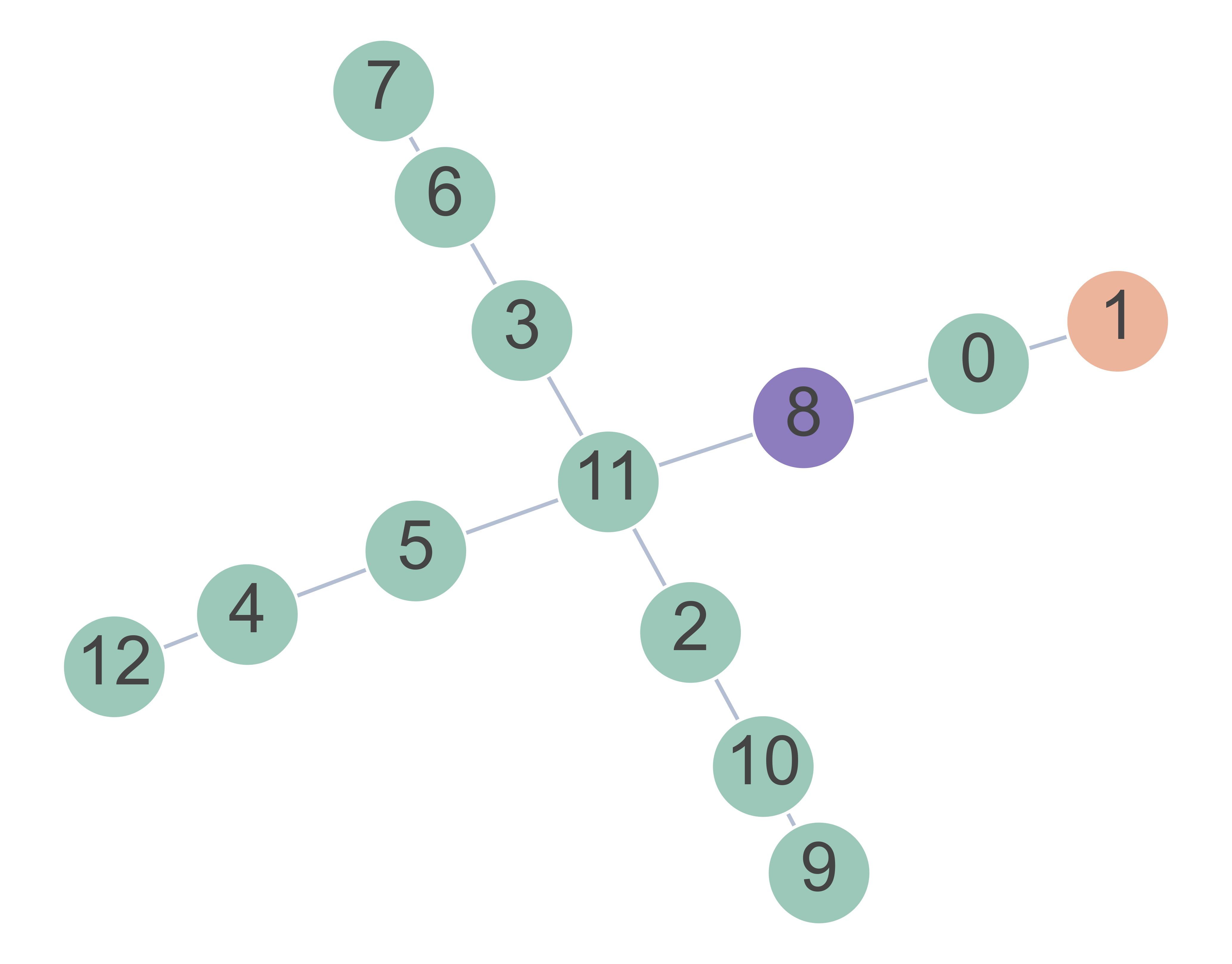}
    \end{minipage}
    \hfill
    \begin{minipage}{0.79\textwidth}
        \centering
        \includegraphics[width=\linewidth]{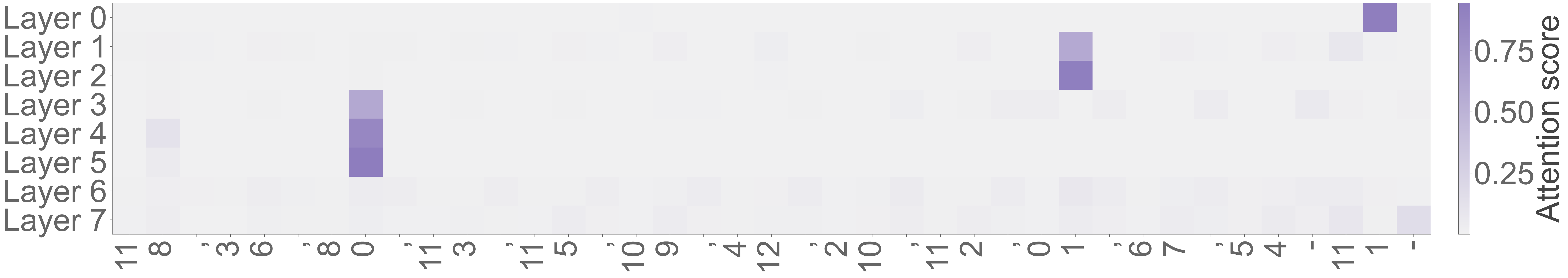}
    \end{minipage}

    \vspace{2em}

    \begin{minipage}{0.19\textwidth}
        \centering
        \includegraphics[width=\linewidth]{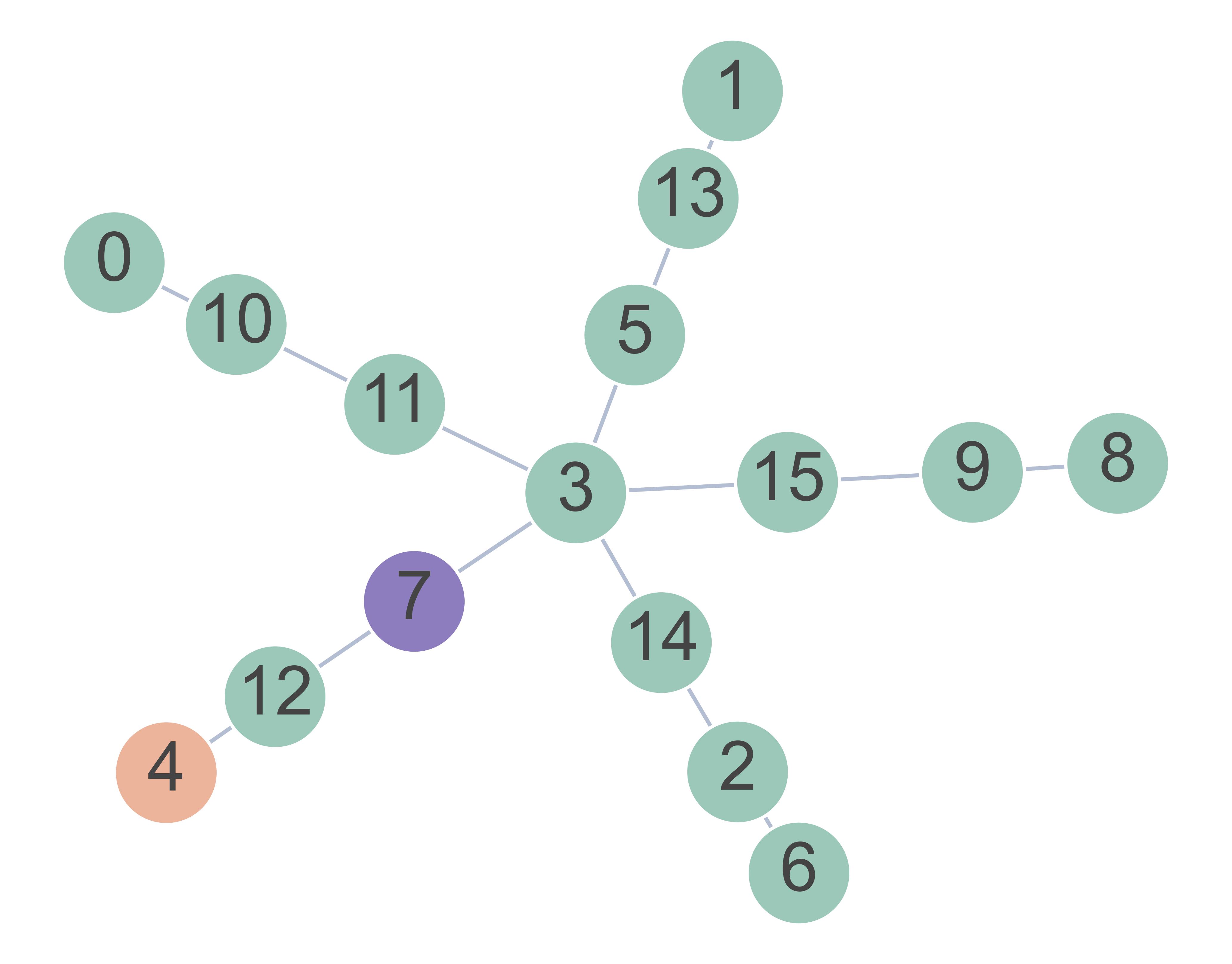}
    \end{minipage}
    \hfill
    \begin{minipage}{0.79\textwidth}
        \centering
        \includegraphics[width=\linewidth]{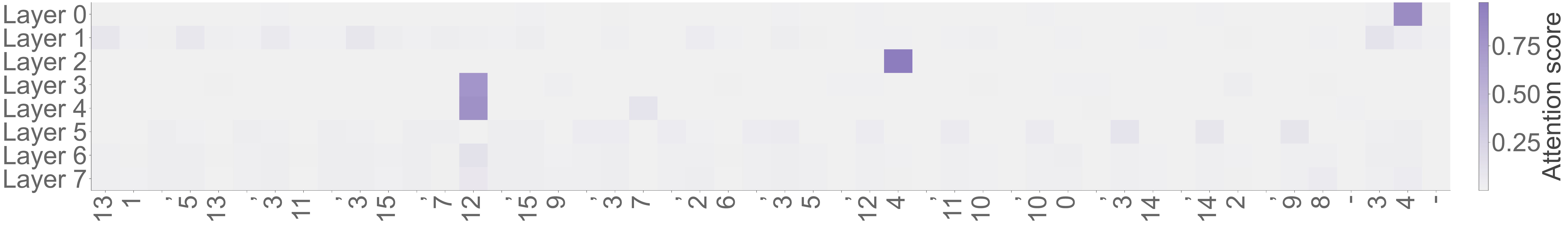}
    \end{minipage}

\caption{Attention visualization for successful configurations ($m=3$, $k \in \{2, 3, 4, 5\}$). For each configuration, the graph structure is shown on the left, where the \textcolor{orange}{orange} node denotes the target and the \textcolor{violet}{violet} node denotes the prediction, and the attention weights from the final token to all input tokens at each transformer layer are shown on the right. As $k$ increases from 2 to 5, the attention progressively concentrates on nodes along the path from $v_{\mathrm{target}}$ to $v_{\mathrm{source}}$, with shallower layers attending to the target and deeper layers tracing back toward the source.}
\label{fig:qualitative-success-configs}
\end{figure*}

\begin{figure*}[t]
    \centering

    \begin{minipage}{0.19\textwidth}
        \centering
        \includegraphics[width=\linewidth]{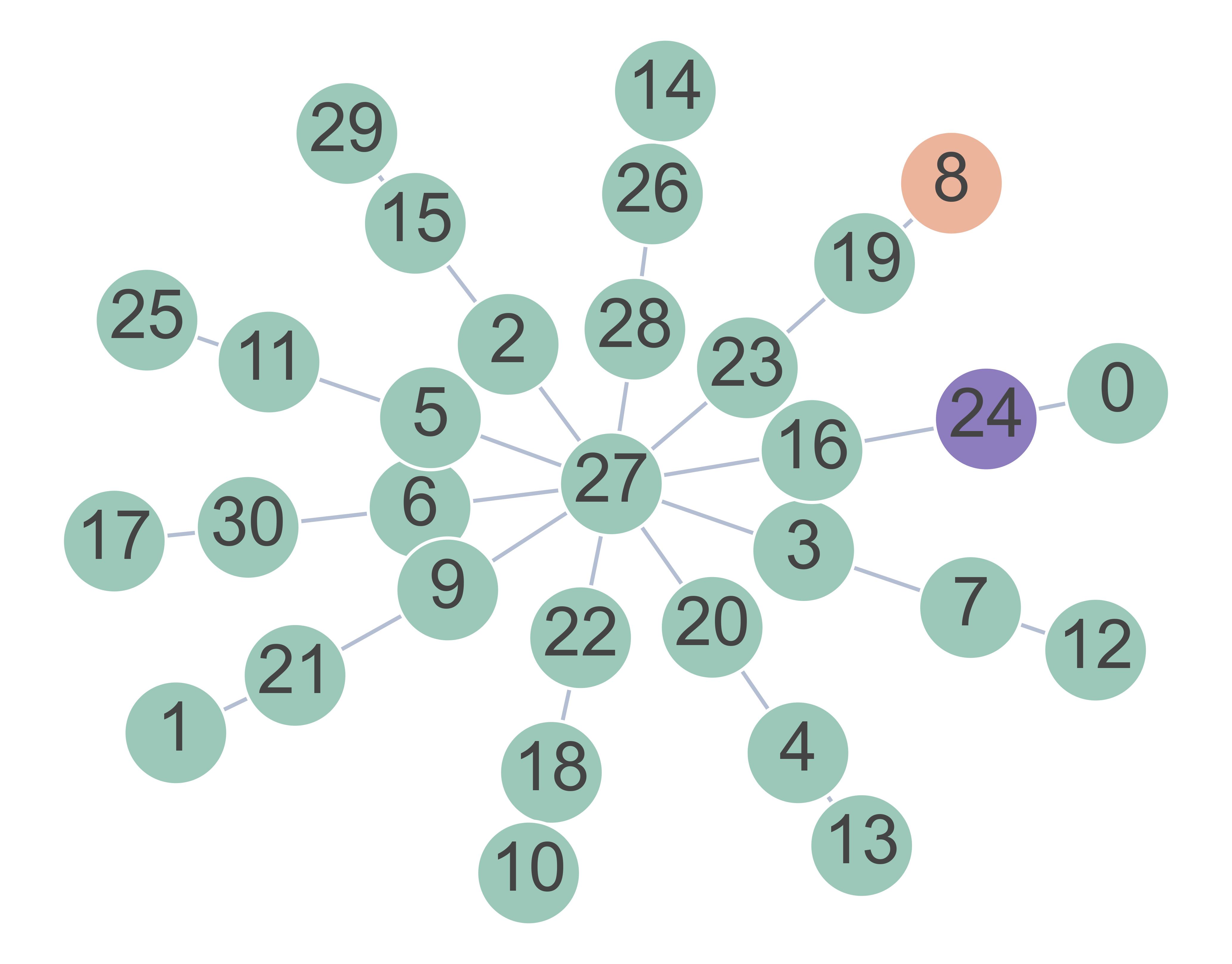}
    \end{minipage}
    \hfill
    \begin{minipage}{0.79\textwidth}
        \centering
        \includegraphics[width=\linewidth]{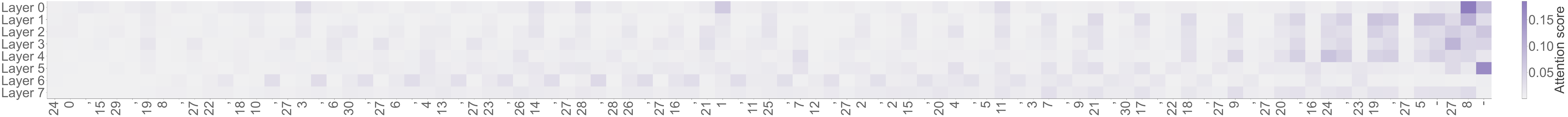}
    \end{minipage}

    \vspace{1.5em}

    \begin{minipage}{0.19\textwidth}
        \centering
        \includegraphics[width=\linewidth]{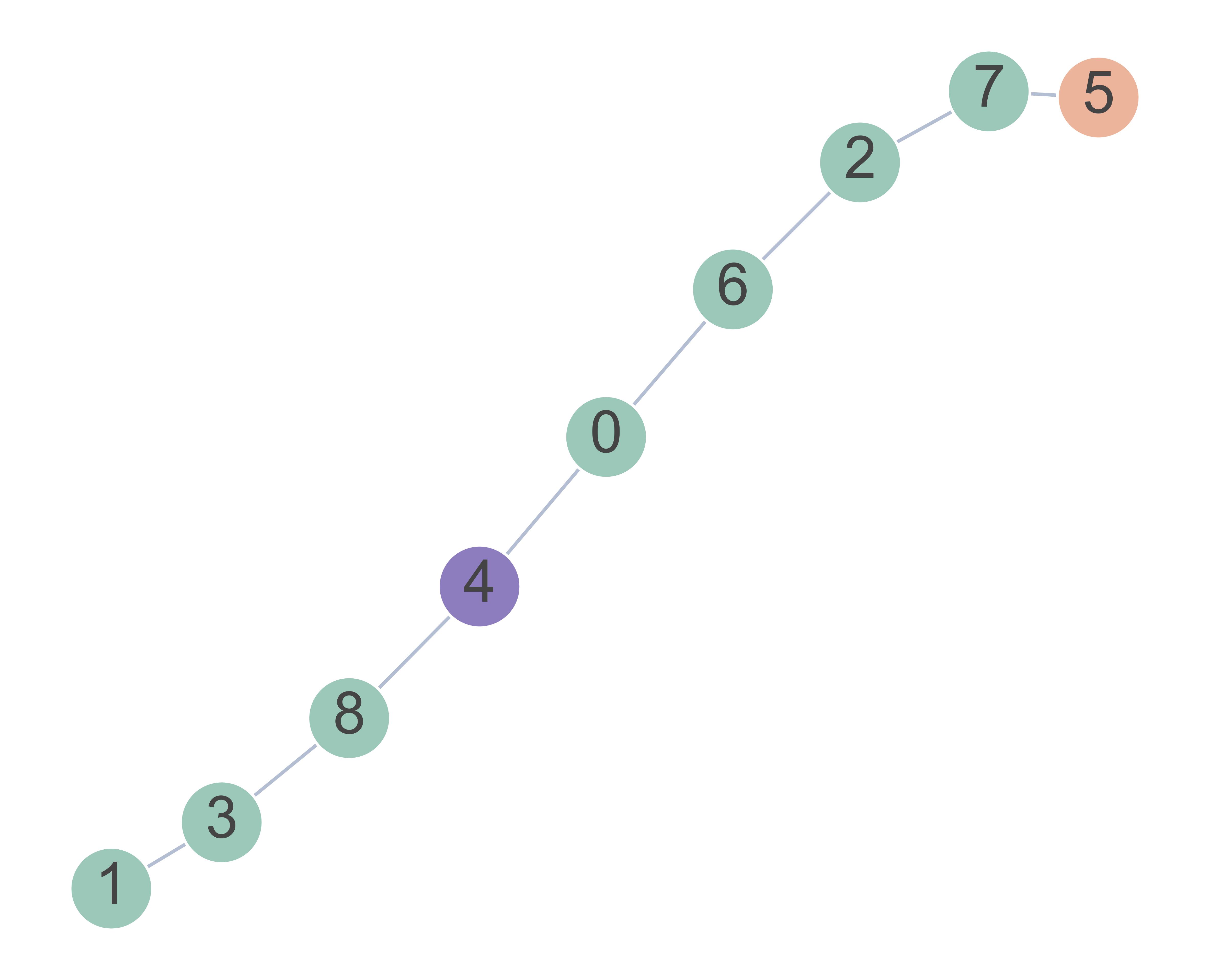}
    \end{minipage}
    \hfill
    \begin{minipage}{0.79\textwidth}
        \centering
        \includegraphics[width=\linewidth]{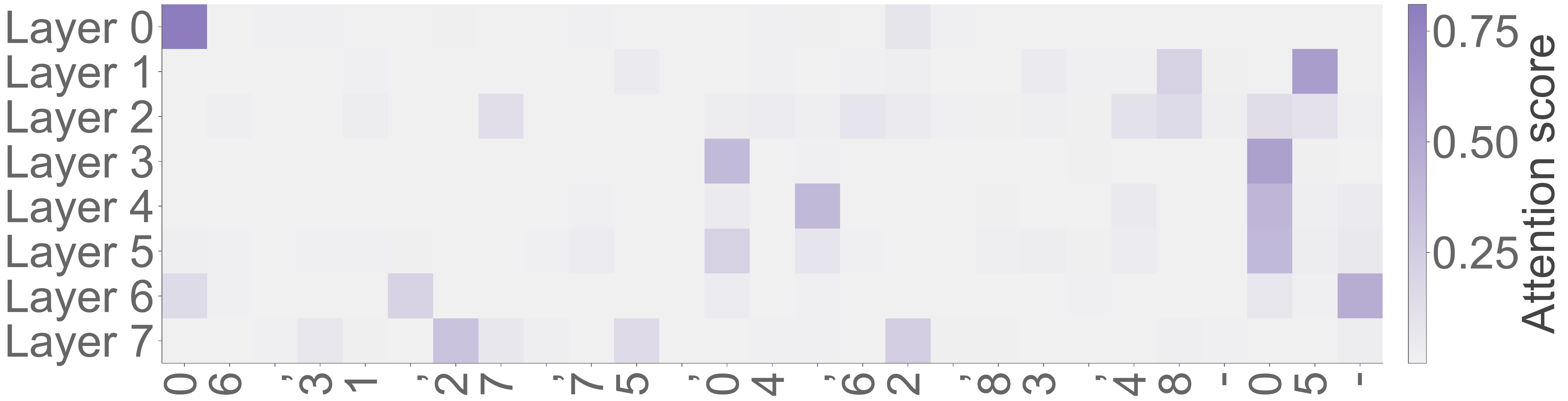}
    \end{minipage}

    \vspace{1.5em}

    \begin{minipage}{0.19\textwidth}
        \centering
        \includegraphics[width=\linewidth]{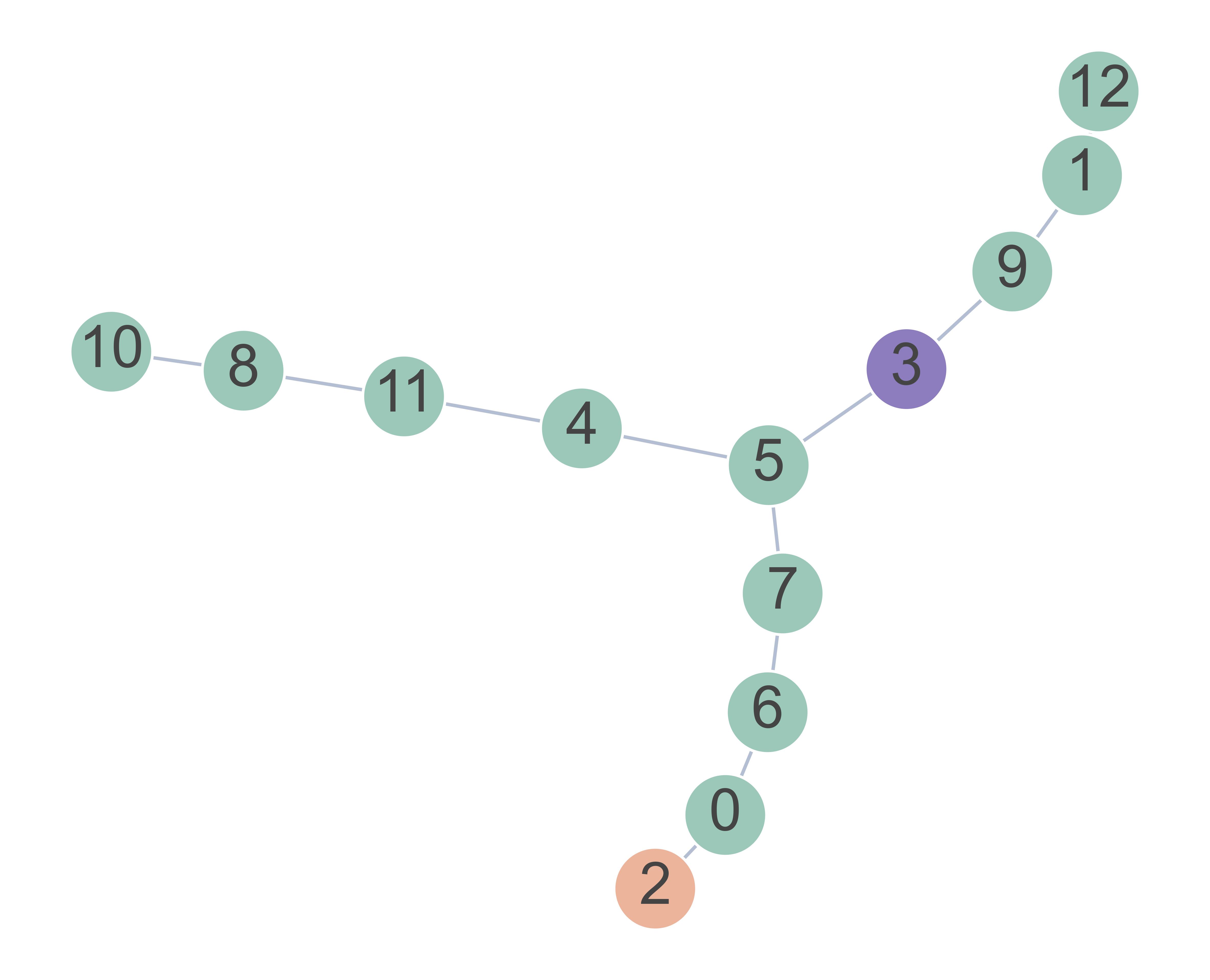}
    \end{minipage}
    \hfill
    \begin{minipage}{0.79\textwidth}
        \centering
        \includegraphics[width=\linewidth]{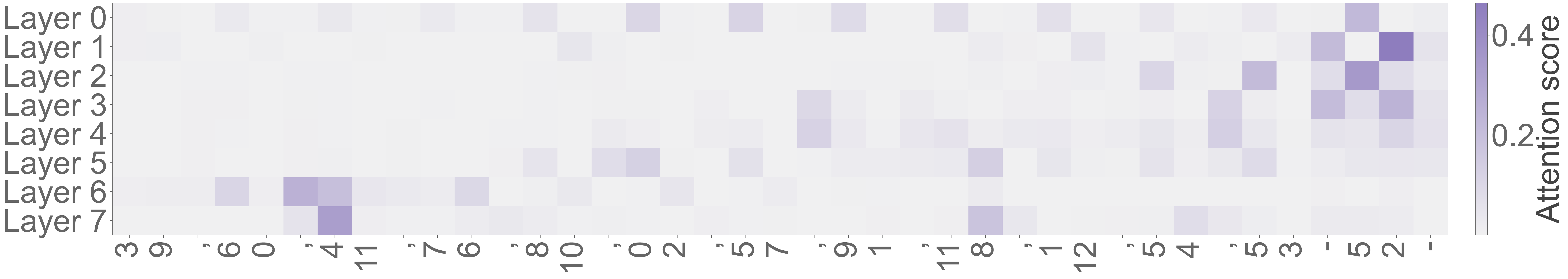}
    \end{minipage}

    \vspace{1.5em}

    \begin{minipage}{0.19\textwidth}
        \centering
        \includegraphics[width=\linewidth]{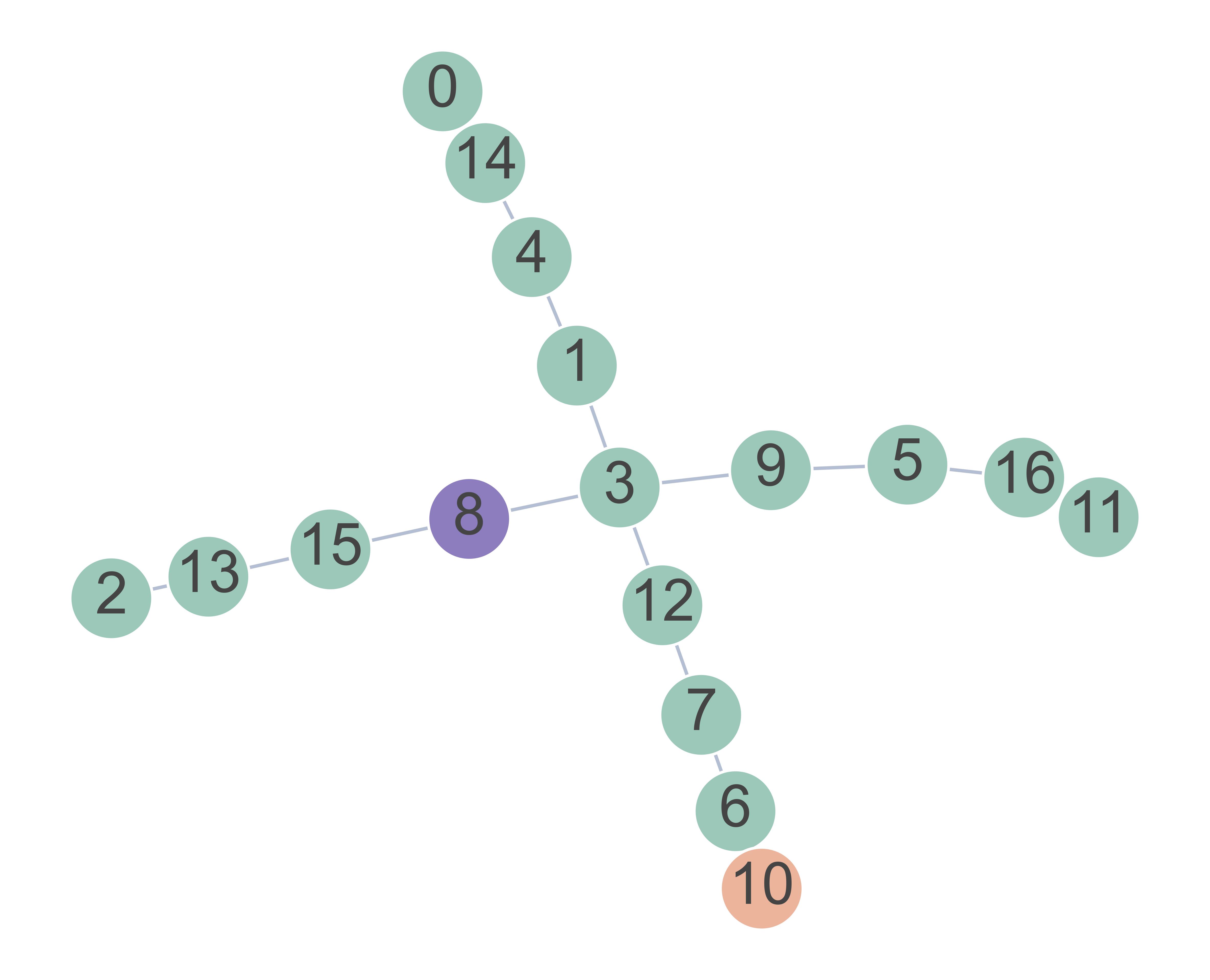}
    \end{minipage}
    \hfill
    \begin{minipage}{0.79\textwidth}
        \centering
        \includegraphics[width=\linewidth]{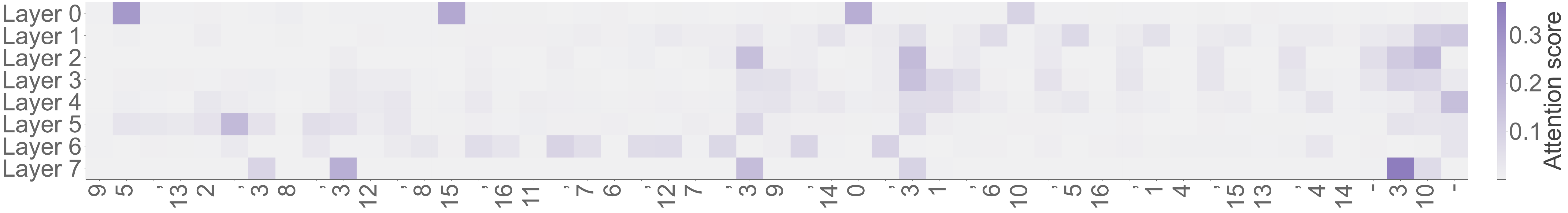}
    \end{minipage}

    \vspace{0.5em}

    \begin{minipage}{0.19\textwidth}
        \centering
        \includegraphics[width=\linewidth]{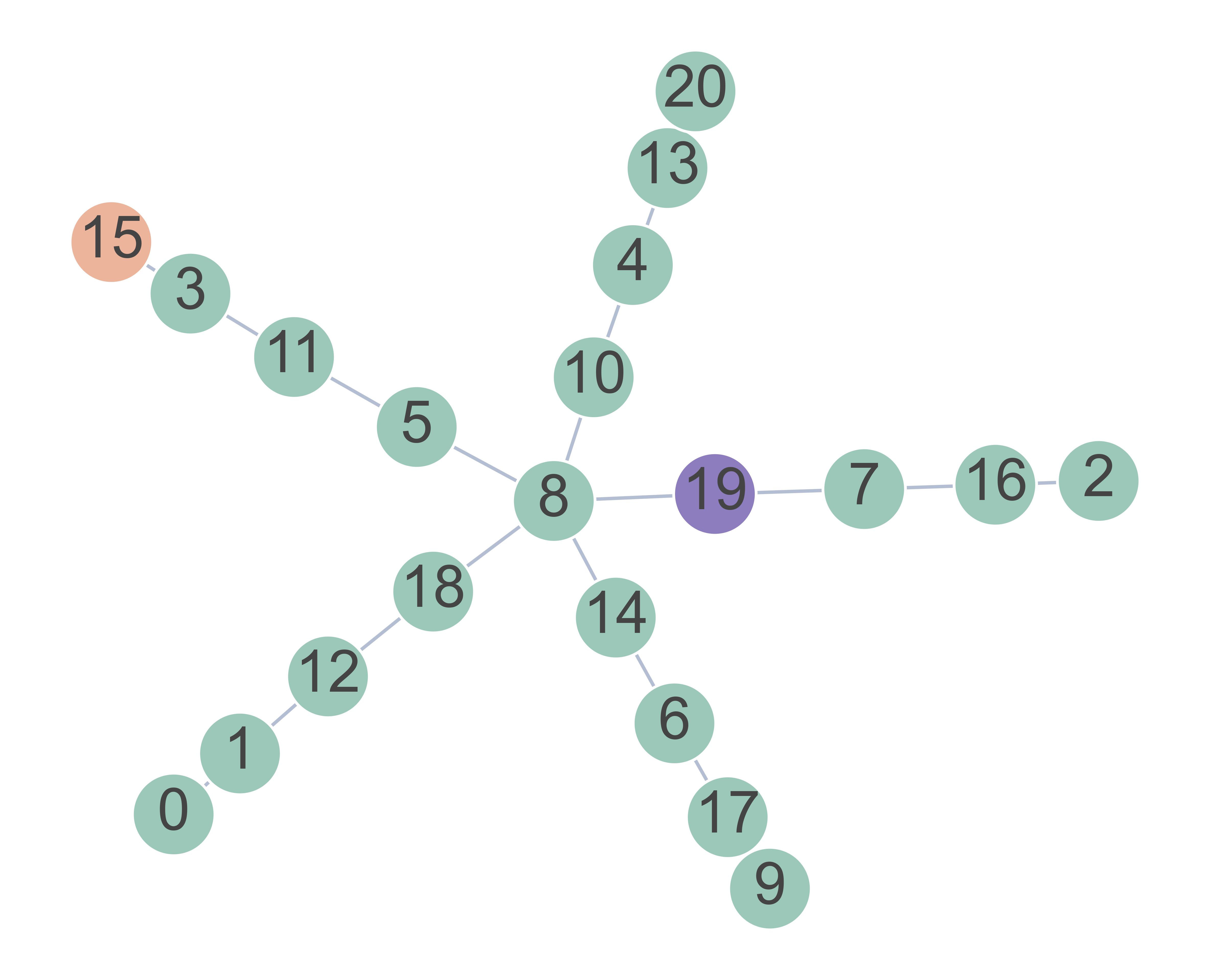}
    \end{minipage}
    \hfill
    \begin{minipage}{0.79\textwidth}
        \centering
        \includegraphics[width=\linewidth]{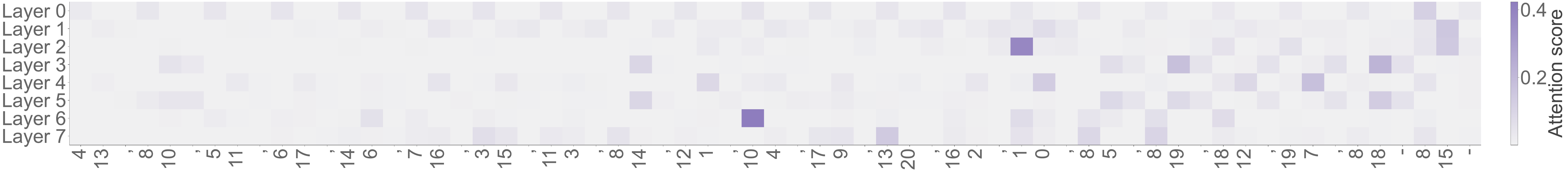}
    \end{minipage}

    \begin{minipage}{0.19\textwidth}
        \centering
        \includegraphics[width=\linewidth]{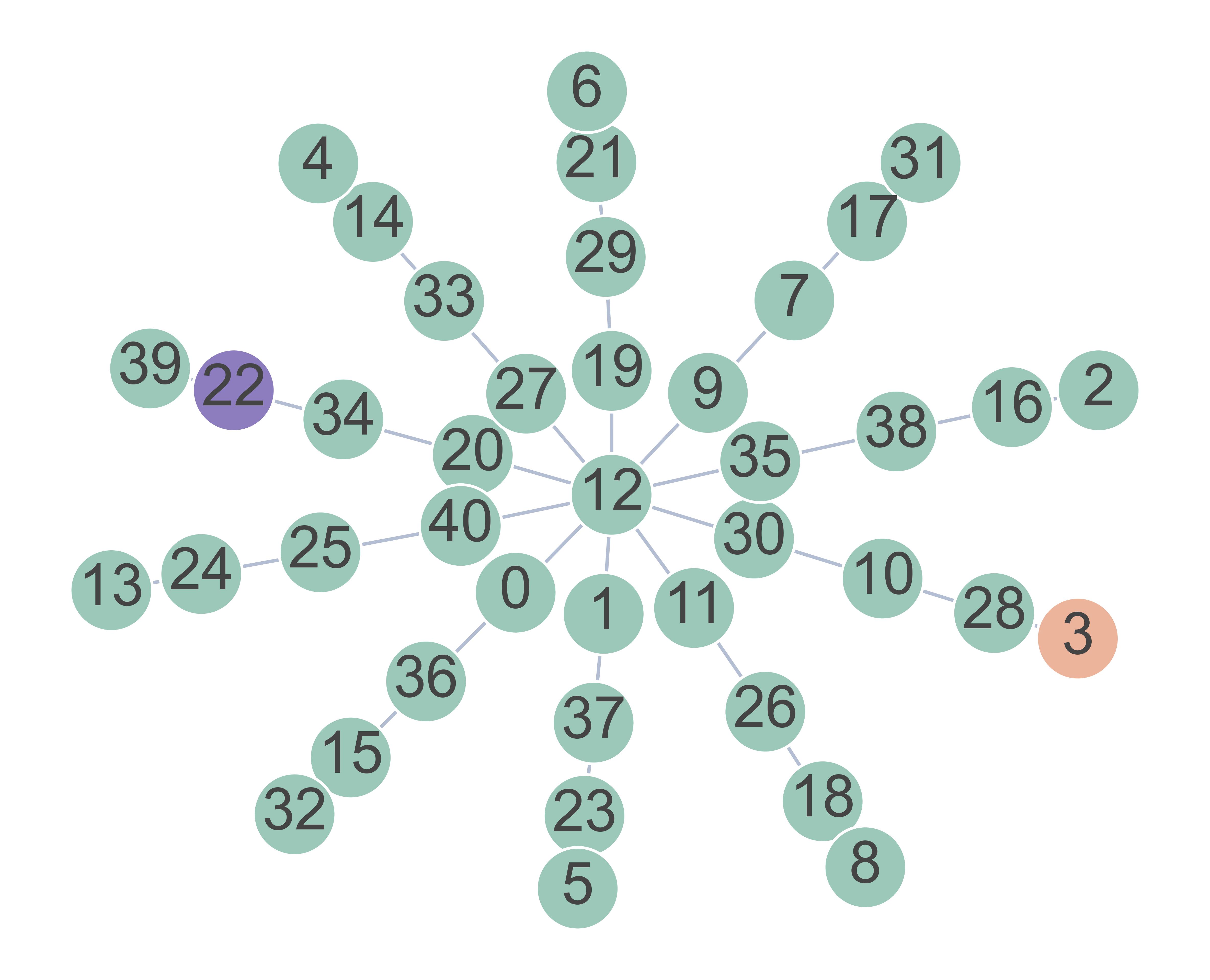}
    \end{minipage}
    \hfill
    \begin{minipage}{0.79\textwidth}
        \centering
        \includegraphics[width=\linewidth]{figs/attention/l4b10/2_attention.pdf}
    \end{minipage}

\caption{Attention visualization for failed configurations ($m=3$ with $k=10$, and $m=4$ with $k \in \{2, 3, 4, 5, 10\}$), following the same convention as Figure~\ref{fig:qualitative-success-configs}: \textcolor{orange}{orange} denotes the target node and \textcolor{violet}{violet} denotes the predicted node. No structured attention pattern emerges across any of these configurations, confirming that the model has not discovered a strategy.}
\label{fig:qualitative-fail-configs}
\end{figure*}

To further investigate the internal strategy of the trained transformer, we visualize the attention weights from the final token to all input tokens at each layer. Specifically, for each test sample, we extract the attention map at every layer and examine how the attention of the model distributes across layers. We provide representative attention visualizations for successful and failed configurations in Figures~\ref{fig:qualitative-success-configs} and~\ref{fig:qualitative-fail-configs}, respectively.

For successful configurations at depth $m=3$, we observe a clear progression as the branch factor increases. When $k=2$, the attention weights exhibit no discernible pattern, consistent with the near-uniform BR of $0.50$ reported in Table~\ref{tab:br-transformer}. This aligns with our earlier analysis: in this special topology, the model does not need to rely on backtracking, as multiple viable strategies exist. As the branch factor increases to $k=4$ and $k=5$, a clear backtracking process emerges in the attention maps. In both cases, the transformer first attends to the target node at shallower layers, then progressively traces the edges from the target back toward the source at deeper layers, mirroring the backtracking strategy described in Appendix~\ref{appendix:strategy_analysis}.

For failed configurations, no such structured pattern is observed, indicating that the model has not converged on a coherent planning strategy. 

\section{LLM Results}

\begin{figure*}[t]
  \centering
  \begin{subfigure}[b]{\textwidth}
    \centering
    \includegraphics[width=0.6\textwidth]{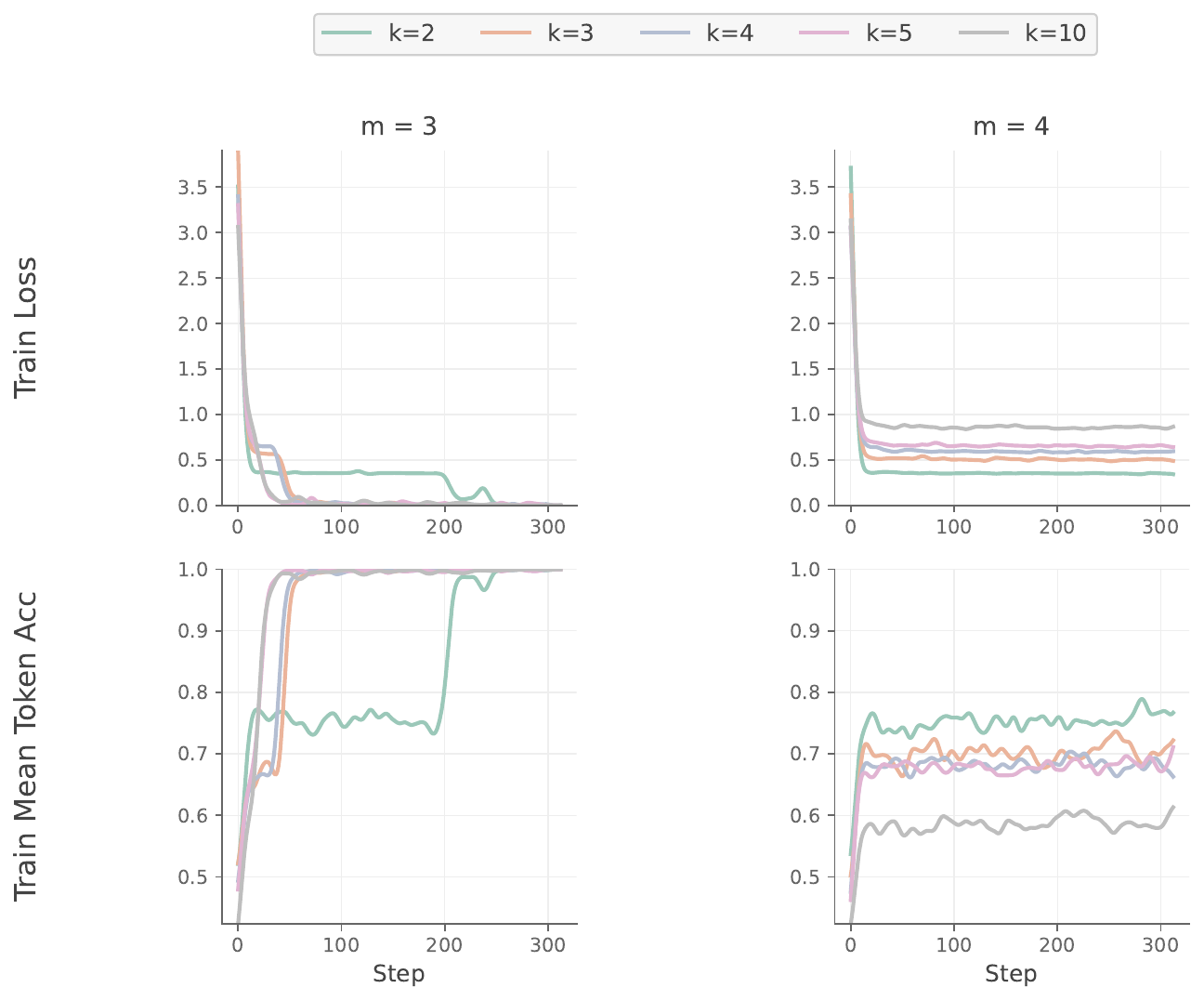}
    \caption{Qwen 2.5-7B: training loss and mean token accuracy across configurations.}
    \label{fig:qwen_25_7b_loss}
  \end{subfigure}
  
  \vspace{5mm}
  
  \begin{subfigure}[b]{\textwidth}
    \centering
    \includegraphics[width=0.7\textwidth]{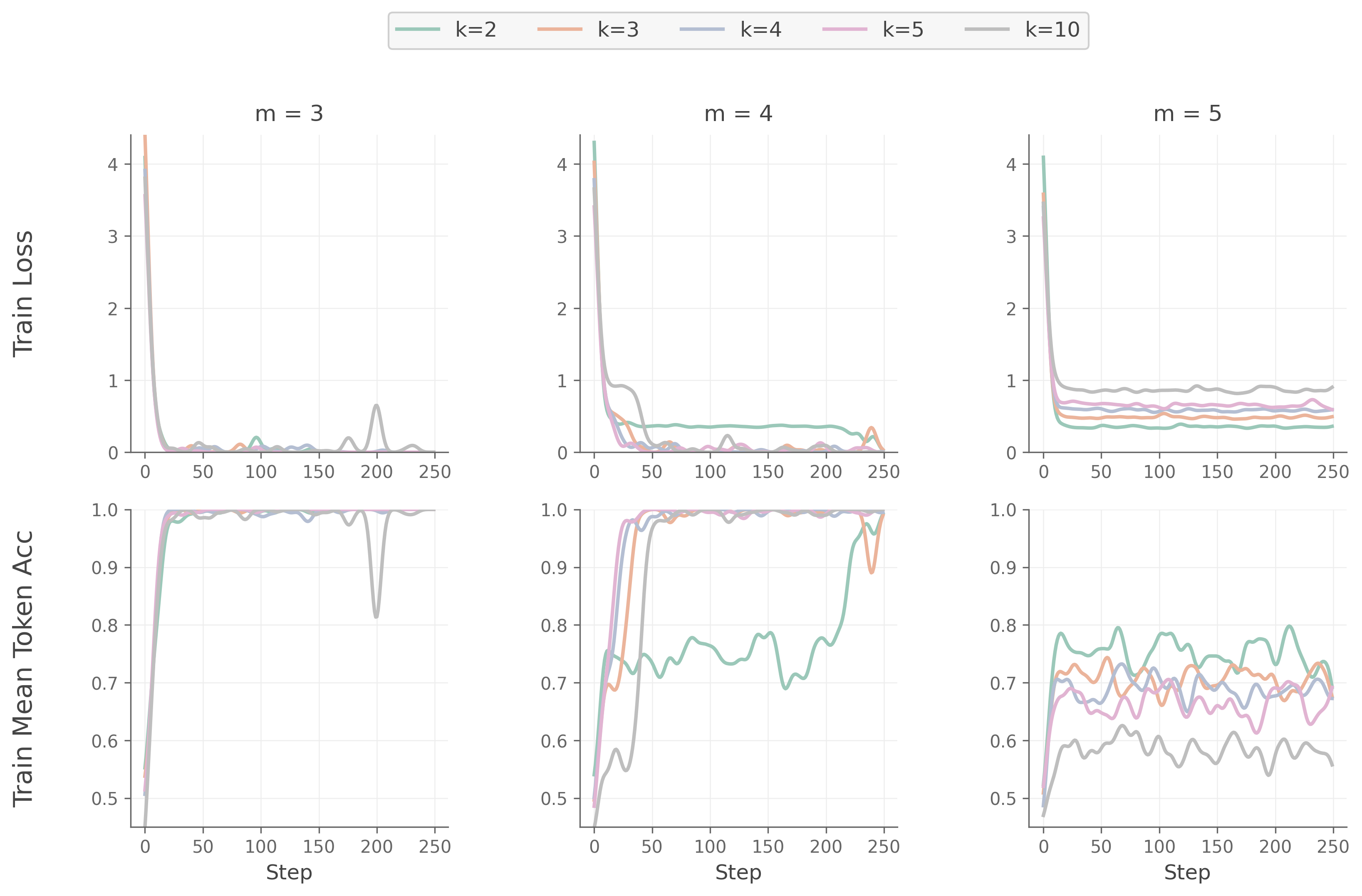}
    \caption{Qwen 2.5-32B: training loss and mean token accuracy across configurations.}
    \label{fig:qwen_25_32b_loss}
  \end{subfigure}
  
  \caption{Training dynamics of Qwen 2.5 models (7B and 32B) across graph configurations. Both models exhibit the same two-stage learning process observed in the from-scratch transformer, with failures occurring exclusively in the second stage.}
  \label{fig:qwen_25_training}
\end{figure*}

\begin{figure*}[t]
  \centering
  \begin{subfigure}[b]{\textwidth}
    \centering
    \includegraphics[width=0.7\textwidth]{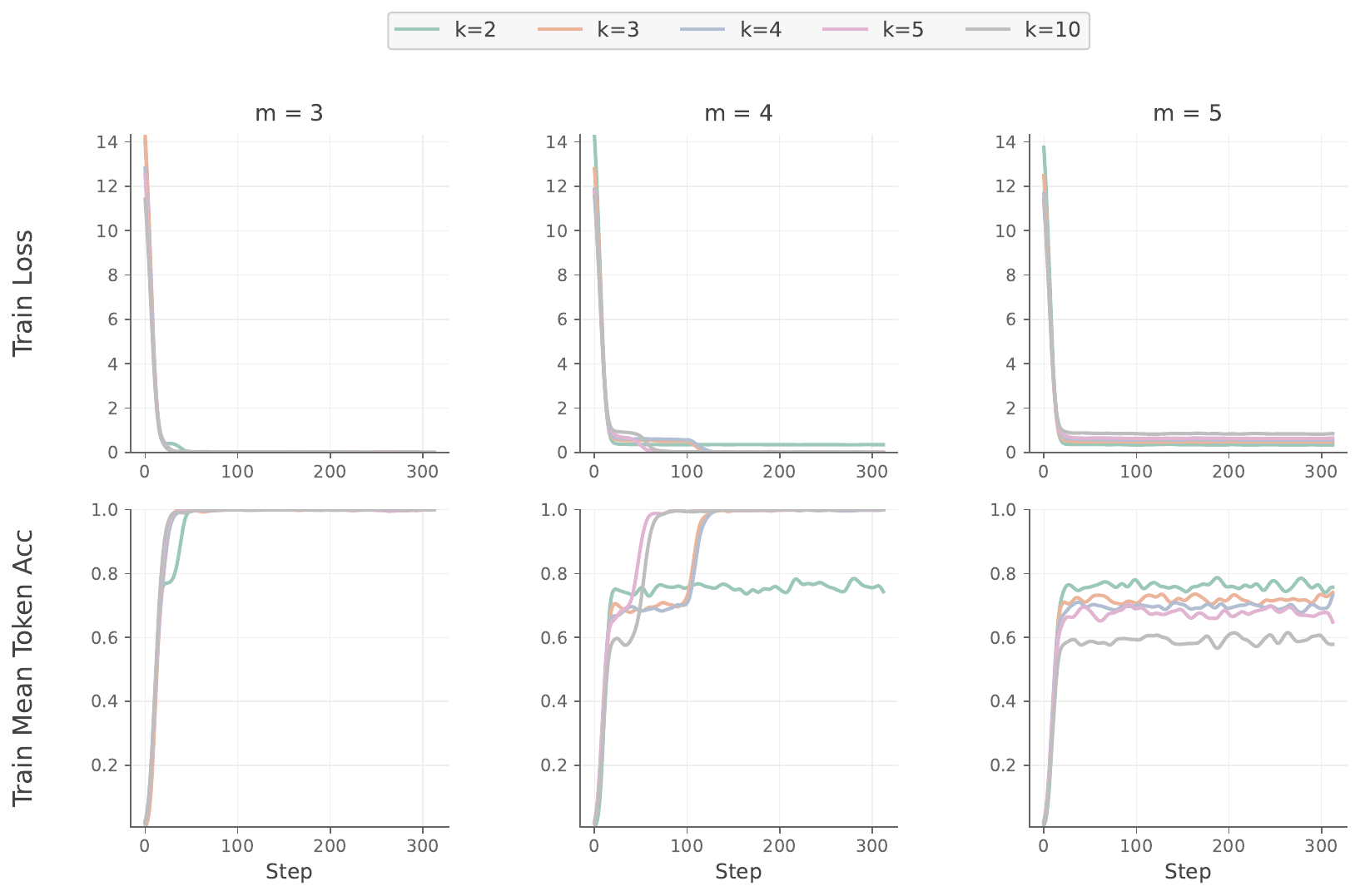}
    \caption{Qwen 3-8B: training loss and mean token accuracy across configurations.}
    \label{fig:qwen_3_8b_loss}
  \end{subfigure}
  
  \vspace{5mm}
  
  \begin{subfigure}[b]{\textwidth}
    \centering
    \includegraphics[width=0.8\textwidth]{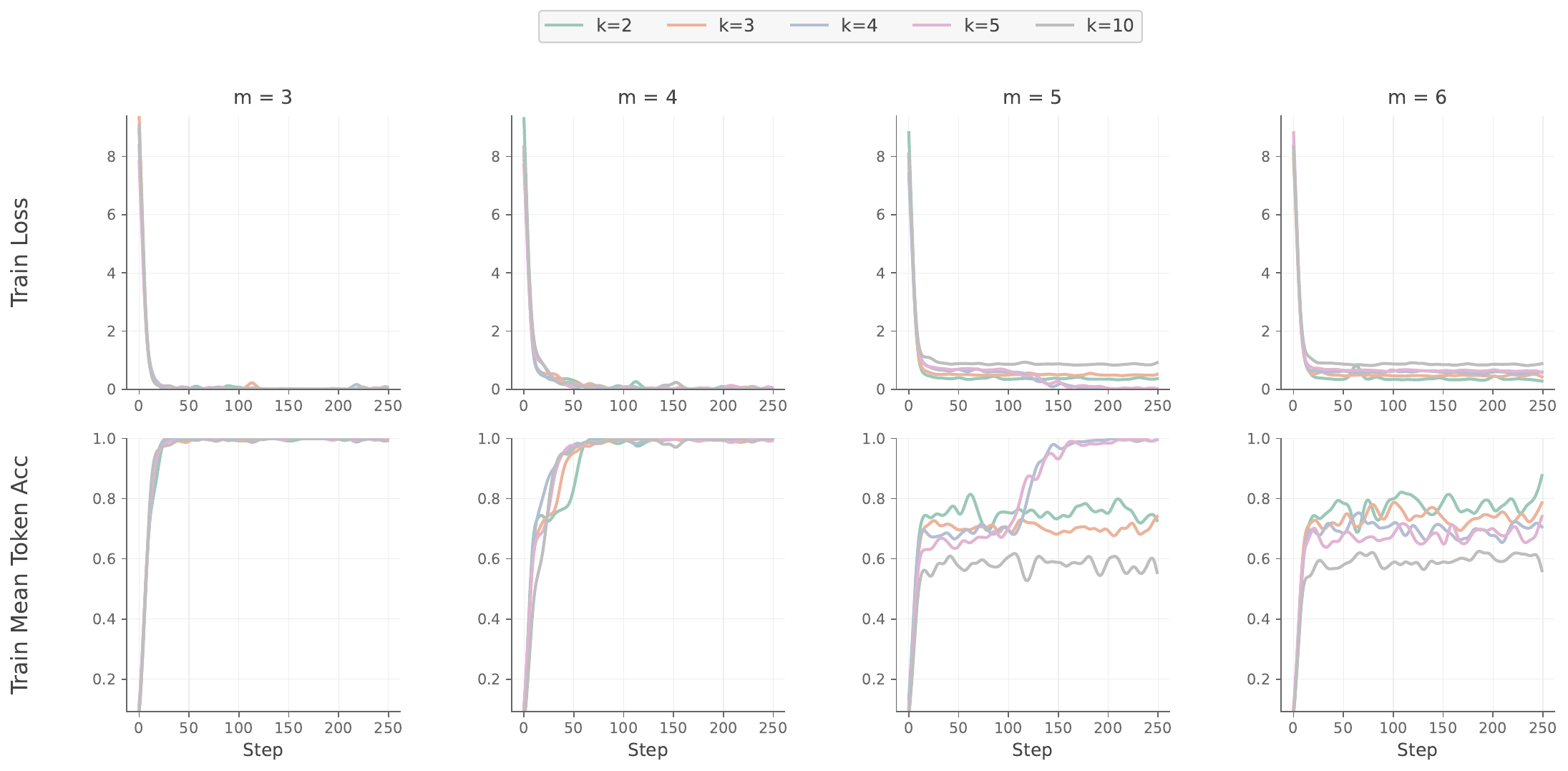}
    \caption{Qwen 3-32B: training loss and mean token accuracy across configurations.}
    \label{fig:qwen_3_32b_loss}
  \end{subfigure}
  
  \caption{Training dynamics of Qwen 3 models (8B and 32B) across graph configurations. The same two-stage pattern persists, confirming that the discovery bottleneck is consistent across model families and scales.}
  \label{fig:qwen_3_training}
\end{figure*}

\subsection{Training Dynamics of LLM}
\label{appendix:llm_training}
As noted in the main text, the training dynamics of fine-tuned LLMs closely mirror the two-stage learning process observed in the from-scratch transformer. Figures~\ref{fig:qwen_25_training} and~\ref{fig:qwen_3_training} present the training loss and mean token accuracy curves for all fine-tuned open-source models across graph configurations. The mean token accuracy during training is computed over all non-masked tokens in the target sequence, including the \texttt{<eos>} token. As a result, the accuracy at the end of the first stage does not equal $\frac{1}{k}$ as in the from-scratch transformer, but the qualitative two-stage pattern remains the same.

\subsection{Zero- and Few-Shot Performance}
\label{appendix:zero_few_shot}
In this subsection, we evaluate LLMs under zero-shot and few-shot prompting (detailed in the Appendix~\ref{app:prompts}) without any task-specific training, to assess whether pre-training alone yields latent planning capabilities. Notably, we exclude Qwen 2.5-7B from this evaluation. We observe that under both zero-shot and few-shot settings, it fails to follow the instruction to directly output $v_{\mathrm{pred}}$. Instead, it persistently generates CoT reasoning, despite explicit instructions in the prompt to avoid doing so.

\paragraph{Zero-shot Performance Reveals Shallow Latent Planning} 
Table~\ref{tab:acc_untrained} shows that all evaluated LLMs exhibit some degree of latent planning ability at shallow depths. At depth $m=3$, even the weakest model (Qwen 3-8B) achieves an LPC of 3, while GPT-4o reaches an LPC of 4 with notably higher skill scores across branch factors. However, performance deteriorates rapidly with increasing depth: at $m=5$, all Qwen models and GPT-4o produce skill scores near or below zero.

Even GPT-5.4, which represents a significant leap in general capability over GPT-4o as reflected by Chatbot Arena rankings \citep{zheng2023judging}, exhibits a qualitatively similar pattern. While it achieves near-perfect skill at $m=3$ (averaging $0.96$) and maintains strong performance at $m=4$ ($0.86$) and $m=5$ ($0.46$), it fails to reject random guessing at $m=6$, yielding a zero-shot LPC of only 5. This extends the planning horizon of GPT-4o by just one step, suggesting that the depth barrier is not easily overcome by scaling alone.

\paragraph{Few-shot Prompting Improves Shallow Performance}
We next evaluate whether few-shot demonstrations can improve the latent planning ability observed in the zero-shot setting. Specifically, we provide three demonstrations for each test instance. 
The demonstrations follow the same graph configuration as the test example but are drawn from disjoint instances that do not appear in the test set. 

As shown in Table~\ref{tab:acc_untrained}, few-shot prompting yields consistent improvements at shallow depths across all models. At $m=3$, the average skill of GPT-4o increases from $0.68$ to $0.93$, and that of Qwen 3-32B improves from $0.23$ to $0.50$. However, these gains diminish rapidly with depth: at $m=5$, all models except GPT-5.4 still collapse to near-random performance. For GPT-5.4, few-shot prompting raises skill from $0.46$ to $0.82$ at $m=5$ and from $0.19$ to $0.63$ at $m=6$, extending its LPC from 5 to 7 (Table~\ref{tab:gpt54_skill_m78}).Yet the model still fails at $m=8$, indicating that few-shot prompting shifts the depth ceiling rather than eliminating it.

\paragraph{The Depth Barrier Is Not Explained by Graph Size}
Across all settings, the consistent pattern of depth-dependent failure cannot be attributed to graph complexity alone. For example, $\mathcal{G}_{(2,6)}$ contains only $13$ nodes, strictly fewer than $\mathcal{G}_{(10,3)}$, yet even GPT-4o fails on the former while succeeding on the latter. This confirms that the bottleneck arises from planning depth rather than graph size.

\section{Bypassing Strategy Discovery via Dense Supervision}
The results from previous sections reveal a persistent bottleneck in strategy discovery when models rely solely on sparse, final-answer supervision. This raises a natural question: rather than attempting to overcome this optimization barrier directly, can we bypass it by explicitly teaching the strategy to solve tasks? In this section, we show that explicitly exposing the intermediate planning structure during training successfully circumvents the bottleneck. Furthermore, we demonstrate that this explicitly learned strategy can be partially distilled back into implicit reasoning.

\subsection{Explicit CoT Solves Discovery}
\label{appendix:cot}

\begin{wraptable}{r}{0.45\columnwidth}
\vspace{-4.6mm}
\centering
\small
\renewcommand{\arraystretch}{0.85}
\setlength{\tabcolsep}{5pt}
\caption{Average skill on star graph tasks across planning depths $m \in \{3,4,5,6,20\}$.}
\label{tab:avg_skill_trained}
\vspace{2mm}

\resizebox{0.45\columnwidth}{!}{%
\begin{tabular}{l ccccc}
\toprule
\multirow{2}{*}{Model} & \multicolumn{5}{c}{Average Skill over $\mathcal{K}$ at Depth $m$} \\
\cmidrule(lr){2-6}
& 3 & 4 & 5 & 6 & 20 \\
\midrule

Transformer & 1.00 & 0.78 & 1.00 & 0.75 & 0.78 \\

Qwen 2.5 & & & & & \\
\phantom{xxx}- 7B  & 1.00 & 1.00 & 1.00 & 1.00 & 1.00 \\
\phantom{xxx}- 32B & 1.00 & 0.99 & 1.00 & 0.99 & 0.99 \\

Qwen 3 & & & & & \\
\phantom{xxx}- 8B  & 1.00 & 1.00 & 1.00 & 1.00 & 1.00 \\
\phantom{xxx}- 32B & 1.00 & 0.99 & 1.00 & 1.00 & 1.00 \\

\bottomrule
\end{tabular}
}
\end{wraptable}

Prior work demonstrates that transformers achieve perfect performance on star graphs when trained to explicitly generate the complete reverse path from $v_{\mathrm{target}}$ to $v_{\mathrm{source}}$ \citep{bachmann2024pitfalls}. We view this success from a different angle: by omitting the final $v_{\mathrm{source}}$, this task essentially reduces to predicting the first-hop node ($v_{\mathrm{ground}}$), with the preceding backtracked nodes naturally serving as an intermediate reasoning process. Motivated by this, we conduct an ablation study formatting these nodes as an explicit CoT rationale. Specifically, we train the model to explicitly backtrack step-by-step from the target until it reaches the ground node, generating the target sequence $\mathbf{y}_{\text{CoT}} = [v_{\mathrm{target}}, \dots, v_{\mathrm{ground}}]$, providing dense supervision that explicitly teaches the path-finding strategy.

\begin{figure*}[t]
  \centering
  \includegraphics[width=0.6\textwidth]{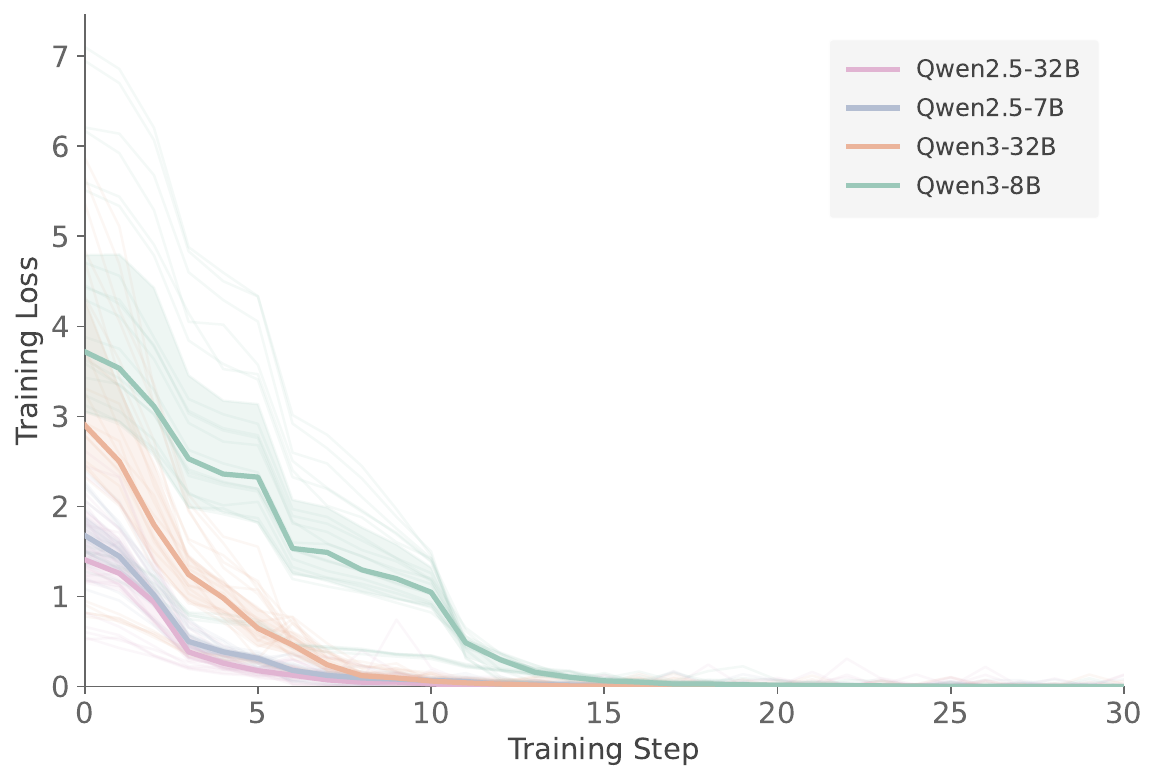}
\caption{\textbf{Training loss under explicit chain-of-thought supervision.} All models are trained with dense supervision that provides the full backtracking trace from $v_{\mathrm{target}}$ to $v_{\mathrm{source}}$. Under this setting, all LLMs converge within approximately 20 training updates across all configurations, confirming that the task itself is not inherently difficult.}
    \label{fig:cot_loss}
\end{figure*}

\begin{table}[t]
\centering
\small
\renewcommand{\arraystretch}{0.85}
\setlength{\tabcolsep}{5pt}
\caption{Skill of ICoT on star graph tasks across branching factors $k \in \mathcal{K}=\{2,3,4,5,10\}$ and planning depths $m \in \{3,4,5,6,7\}$.
\win{Blue-highlighted bold text} indicates configurations where the model successfully rejects random guessing, while \fail{gray text} indicates failure.}
\label{tab:icot_skill}
\vspace{2mm}
\resizebox{0.48\columnwidth}{!}{%
\begin{tabular}{c ccccc}
\toprule
\multirow{2}{*}{$m$} & \multicolumn{5}{c}{$k \in \mathcal{K}$} \\
\cmidrule(lr){2-6}
& 2 & 3 & 4 & 5 & 10 \\
\midrule
3 & \win{1.00} & \win{1.00} & \win{1.00} & \win{1.00} & \fail{-0.01} \\
4 & \win{0.99} & \fail{0.01} & \fail{0.00} & \fail{-0.01} & \fail{-0.11} \\
5 & \win{1.00} & \fail{0.02} & \fail{-0.33} & \fail{0.00} & \fail{-0.11} \\
6 & \win{1.00} & \fail{-0.01} & \fail{-0.33} & \fail{-0.25} & \fail{-0.11} \\
7 & \fail{-1.00} & \fail{-0.50} & \fail{-0.33} & \fail{-0.25} & \fail{-0.11} \\
\bottomrule
\end{tabular}
}
\end{table}

Under this setting, all LLMs rapidly converge with 20 steps (Figure~\ref{fig:cot_loss}) and maintain a near-perfect average skill score across all combinations of planning depths $m \in \{3, 4, 5, 6\}$ and branch factors $k \in \mathcal{K}$, and even at an extended depth of $m=20$ (Table~\ref{tab:avg_skill_trained}). The from-scratch transformer also shows significant improvement, though its average skill drops to $0.60$ at certain depths. A detailed breakdown reveals that while this small model achieves $100\%$ accuracy on most configurations, it occasionally fails ($0\%$ accuracy within 500 epochs) on graphs with larger branch factors, specifically $k=10$ (for $m \in \{4, 20\}$) and $k=5$ (for $m=6$). 

Taken together, these results demonstrate that the previously observed failures at deeper configurations (e.g., $m=6$) are not absolute architectural limits. Instead, they represent a strict discovery bottleneck specific to sparse, final-answer supervision. By leveraging the output context window to externalize intermediate planning states, dense CoT supervision effectively decouples the challenge of strategy discovery from strategy execution. While the small transformer eventually encounters capacity limitations when the graph becomes complex, the LLMs can readily learn to execute deep multi-step planning once the strategy is made explicit.

\subsection{Mitigating Implicit Planning Limitations with ICoT Training}
\label{appendix:icot}
While explicit CoT supervision significantly simplifies learning, it requires the model to generate the full reasoning trace during inference. We therefore examine whether the externalized planning process in $\mathbf{y}_{\text{CoT}}$ can be compressed into the model's internal representations, allowing the same computation to be carried out implicitly.

To do so, we adopt the ICoT training framework \citep{deng2023implicit,deng2024explicit}, which gradually internalizes reasoning by progressively removing tokens from the full response. Specifically, let
\[
\mathbf{y}_{\text{CoT}} = [z_1, \dots, z_m]
= [v_{\mathrm{target}}, \dots, v_{\mathrm{ground}}].
\]
We begin with the explicit CoT objective (stage $s=0$), where the model is trained to predict the full sequence $\mathbf{y}_{\text{CoT}}$. At each subsequent stage $s$, we remove the first $s$ tokens from the original sequence and train the model on the truncated target
\[
\mathbf{y}^{(s)} = [z_{s+1}, \dots, z_m].
\]
As $s$ increases, the supervision sequence evolves as
\[
[z_1, \dots, z_m]
\rightarrow
[z_2, \dots, z_m]
\rightarrow \cdots \rightarrow
[z_m = v_{\mathrm{ground}}],
\]
forcing the model to compute these missing steps within its hidden states.

As Table~\ref{tab:icot_skill} illustrates, ICoT successfully enables the 1.6M model to bypass the discovery bottleneck on simpler graphs, achieving perfect latent planning (up to $m=6$ at $k=2$). However, as the structural complexity of the graph increases—either through greater depth or wider branching—the performance rapidly collapses, demonstrating that while explicit supervision via ICoT effectively eliminates the discovery bottleneck, the intrinsic representational capacity of the small model ultimately becomes the primary limitation.

\section{Hypothesis Testing and Critical Threshold}
\label{appendix:skill}

We use hypothesis testing to determine whether a model significantly outperforms the random baseline. The null hypothesis $H_0$ posits that the model guesses randomly, with a success probability of $\frac{1}{k}$. The one-sided alternative $H_1$ is that the true success probability exceeds $\frac{1}{k}$, or equivalently, that the skill is above $0$. Let $\hat{n}$ denote the number of correct predictions among $\hat{N}$ test samples, so that $\mathrm{Acc}(\pi_{\theta}, k, m) = \frac{\hat{n}}{\hat{N}}$. We reject $H_0$ at significance level $\alpha$ when the upper-tail binomial probability satisfies:
\begin{equation*}
    \sum_{i = \hat{n}}^{\hat{N}} \binom{\hat{N}}{i} \biggl(\frac{1}{k}\biggr)^i \biggl(1 - \frac{1}{k}\biggr)^{\hat{N} - i} \leq \alpha.
\end{equation*}
The minimum empirical accuracy required to reject $H_0$ for different test sample sizes $\hat{N}$, branch factors $k$, and significance levels $\alpha$ are reported in Table~\ref{tab:acc-skill-grid}. We further define the critical skill threshold $\tau_{\mathrm{crit}}(k, \hat{N}, \alpha)$ as the corresponding minimum skill, whose values are also included in the table.

\section{Prompt Templates}
\label{app:prompts}
In this section, we detail the prompt templates utilized for LLM training and evaluation.
\subsection{Prompts For Fine-tuned LLMs}
\begin{table*}[t!]
\centering
\small
\renewcommand{\arraystretch}{1}
\setlength{\tabcolsep}{5pt}
\caption{
Minimal accuracy and critical skill thresholds for rejecting the random-guessing null hypothesis at significance level $\alpha = 10^{-5}$. A model is considered to perform significantly better than random guessing when its empirical accuracy (or skill) exceeds the corresponding threshold.
}
\label{tab:acc-skill-grid}
\begin{adjustbox}{max width=\textwidth}
\begin{tabular}{l ccccc|ccccc}
\toprule
& \multicolumn{5}{c}{$\hat{N}=100$}
& \multicolumn{5}{c}{$\hat{N}=2048$} \\
\cmidrule(lr){2-6}\cmidrule(lr){7-11}
& 2 & 3 & 4 & 5 & 10
& 2 & 3 & 4 & 5 & 10 \\
\midrule
Minimal Accuracy
& .720 & .550 & .460 & .390 & .260
& .547 & .379 & .292 & .239 & .130 \\
Critical Skill
& .440 & .325 & .280 & .238 & .178
& .095 & .068 & .056 & .048 & .033 \\
\bottomrule
\end{tabular}
\end{adjustbox}
\end{table*}
The prompt templates in this subsection are used for fine-tuning LLMs. For all LLMs, we use the default system prompt and modify only the user prompt. For fine-tuned models, we also use the same prompt format during evaluation to ensure consistency between training and evaluation.

\clearpage
\noindent\begin{minipage}{0.9\textwidth}
\begin{tcolorbox}[title=Latent Planning (Direct-Answer) Prompt Template, colback=white, colframe=black!60]
\footnotesize

\textbf{[User Prompt]} \\
You are given an undirected graph. The graph is described as a list of edges. \\
Each edge is written as two node indices separated by a space, and different edges are separated by commas.\\
Graph: \{\texttt{graph}\}\\
Source: \{\texttt{source}\}\\
Target: \{\texttt{target}\}\\
Your task is to only return the first hop node after the Source node on the shortest path toward the Target node.\\
Return an integer only.
\end{tcolorbox}
\end{minipage}

\noindent\begin{minipage}{0.9\textwidth}
\begin{tcolorbox}[title=Backtracking (CoT) Prompt Template, colback=white, colframe=black!60]
\footnotesize

\textbf{[User Prompt]} \\
You are given an undirected graph. The graph is described as a list of edges. \\
Each edge is written as two node indices separated by a space, and different edges are separated by commas.\\
Graph: \{\texttt{graph}\}\\
Source: \{\texttt{source}\}\\
Target: \{\texttt{target}\}\\
Your task is to only return the first hop node after the Source node on the shortest path toward the Target node.\\
You can find this first hop by backtracking from the Target node along a shortest path toward the Source node until you reach a direct neighbor of the Source node. \\
Return the backtracking nodes as integers separated by single spaces.\\
The last integer must be the first hop node.
\end{tcolorbox}
\end{minipage}

\subsection{Prompts For Pre-trained LLMs}
In preliminary experiments, we found that directly using the same prompt templates as those used for fine-tuning often led these models to violate the output-format instructions by generating additional explanations or reasoning steps. To reduce such instruction-following failures, we use more explicit prompt templates for these models, with stronger formatting constraints and clearer output requirements in both system and user prompts.

\noindent\begin{minipage}{0.9\textwidth}
\begin{tcolorbox}[title=Zero-shot Prompt Template, colback=white, colframe=black!60]
\footnotesize
\textbf{[System Prompt]} \\
You are a precise assistant. You must follow the instructions exactly and output only what is explicitly requested. Do not provide explanations or intermediate reasoning.

\vspace{6pt}
\textbf{[User Prompt]} \\
You are given an undirected graph. The graph is described as a list of edges.
Each edge is written as two node indices separated by a space, and different edges are separated by commas.\\
Graph: \{\texttt{graph}\}\\
Source: \{\texttt{source}\}\\
Target: \{\texttt{target}\}\\
Your task is to find the first hop node after the Source node on the shortest path toward the Target node.\\
Return ONLY the first hop node after the Source on this path.\\
Do NOT provide explanations, reasoning, or intermediate steps.\\
Output ONLY the final answer enclosed within \texttt{<}answer\texttt{>} and \texttt{<}/answer\texttt{>} tags. \\
For example: \texttt{<}answer\texttt{>}3\texttt{<}/answer\texttt{>}
\end{tcolorbox}
\end{minipage}

\noindent\begin{minipage}{0.9\textwidth}
\begin{tcolorbox}[title=Few-shot Prompt Template, colback=white, colframe=black!60]
\footnotesize
\textbf{[System Prompt]} \\
You are a precise assistant. You must follow the instructions exactly and output only what is explicitly requested. Do not provide explanations or intermediate reasoning.

\vspace{6pt}
\textbf{[User Prompt]} \\
You are given an undirected graph. The graph is described as a list of edges.
Each edge is written as two node indices separated by a space, and different edges are separated by commas.\\
Your task is to find the first hop node after the Source node on the shortest path toward the Target node.\\
Return ONLY the first hop node after the Source on this path.\\
Do NOT provide explanations, reasoning, or intermediate steps.\\
Output ONLY the final answer enclosed within \texttt{<}answer\texttt{>} and \texttt{<}/answer\texttt{>} tags. \\
Below are some examples and their solutions.\\
At the end, you will be given another question that you will need to solve. Make sure you follow the same format.\\

\vspace{6pt}
\textbf{[Example 1]}\\
Graph: \{\texttt{graph}\}\\
Source: \{\texttt{source}\}\\
Target: \{\texttt{target}\}\\
\texttt{<}answer\texttt{>}\{\texttt{answer}\}\texttt{<}/answer\texttt{>}

\vspace{6pt}
\textbf{[Example 2]}\\
Graph: \{\texttt{graph}\}\\
Source: \{\texttt{source}\}\\
Target: \{\texttt{target}\}\\
\texttt{<}answer\texttt{>}\{\texttt{answer}\}\texttt{<}/answer\texttt{>}

\vspace{6pt}
\textbf{[Example 3]}\\
Graph: \{\texttt{graph}\}\\
Source: \{\texttt{source}\}\\
Target: \{\texttt{target}\}\\
\texttt{<}answer\texttt{>}\{\texttt{answer}\}\texttt{<}/answer\texttt{>}

\vspace{6pt}
Now solve the following problem.\\
Graph: \{\texttt{graph}\}\\
Source: \{\texttt{source}\}\\
Target: \{\texttt{target}\}
\end{tcolorbox}
\end{minipage}

\end{document}

%% file: secs/01_methods.tex
\section{Problem Formulation}
Our goal is to study whether models can discover and execute latent planning strategies when planning is strictly required for task success, yet no planning supervision is provided. To this end, we first identify the properties a task must satisfy to isolate latent planning. Specifically, the task must require multi-step lookahead for correct decision making, while ruling out local cues or simple heuristics that could bypass planning. Moreover, task difficulty should be controllable, such that increasing difficulty directly increases the number of planning steps required. All models are trained via standard next-token prediction with cross-entropy loss on the final answer, without supervision on intermediate reasoning steps (see Appendix~\ref{appendix:ntp} for details). We train with next-token prediction because reinforcement learning settings typically permit reasoning to be externalized through intermediate tokens, making latent planning difficult to enforce.

\subsection{Star Graph}
Most complex real-world reasoning and planning tasks can be abstracted to graphs of entities and relations \citep{battaglia2018relational}, making graph path-finding a natural domain for our study. Crucially, it meets our requirements: the correct next-hop choice is not determined by local graph structure and can only be deduced by reasoning over nodes several steps away. We adopt star graphs~\citep{bachmann2024pitfalls} as the underlying topology, offering a clear and controllable environment. Unlike benchmarks such as Blocksworld \citep{valmeekam2023on}, where nontrivial performance can arise from shallow heuristics, this setting ensures that accuracy tightly constrains latent planning depth. 

Given a source node $v_{\mathrm{source}}$, we construct a star graph $G_{(k, m)} \in \mathcal{G}_{(k,m)}$ with $k$ branches, each a simple path extending from $v_{\mathrm{source}}$ through $m$ additional nodes. We then uniformly sample one branch and designate its \textbf{end} node as the target $v_{\mathrm{target}}$.  While the model could theoretically be tasked with generating the complete path from $v_{\mathrm{source}}$ to $v_{\mathrm{target}}$ , identifying the first node on the correct branch essentially trivializes the rest of the path. Thus, the task is strictly limited to finding the correct first node $v_{\mathrm{ground}}$ (Figure~\ref{fig:lpc_bar}, right).

As all branches have identical length and structure, the correct next-hop cannot be determined using local heuristics based on length or degree. Solving this task requires the model to propagate information across the entire trace between $v_{\mathrm{source}}$ and $v_{\mathrm{target}}$ through a sequence of latent computations. This symmetry also provides precise control over the number of internal planning steps required. A systematic analysis of possible solving strategies and their computational requirements is provided in Appendix~\ref{appendix:strategy_analysis}.

\subsection{Metrics}
\paragraph{Latent Planning Capacity}

Since tasks with different branch factors $k$ yield different random baselines, raw \textit{accuracy} does not allow fair comparison across configurations. We therefore first normalize accuracy into an \textit{empirical skill} score $\text{Skill}(\pi_{\theta}, k, m)$, where a value of $1$ indicates perfect performance and $0$ corresponds to random guessing (detailed in Appendix~\ref{appendix:metric}). Building on this, we define the \textit{latent planning capacity} (LPC) to capture whether a model exhibits any statistically significant evidence of planning at a given depth $m$.

Concretely, we evaluate each model on a set of branch factors $\mathcal{K} = \{2, 3, 4, 5, 10\}$ and test whether its performance is statistically distinguishable from random guessing when $\alpha = 10^{-5}$. A model $\pi_\theta$ is said to possess latent planning capability at depth $m$ if its empirical skill exceeds the critical threshold $\tau_{\mathrm{crit}}$ for at least one $k \in \mathcal{K}$:
\[
\mathrm{LPC}(\pi_\theta, m)
=
\begin{cases}
1, & \max_{k \in \mathcal{K}} \left[ \mathrm{Skill}(\pi_\theta, k, m) - \tau_{\mathrm{crit}}(k, \hat{N}, \alpha) \right] \ge 0, \\
0, & \text{otherwise},
\end{cases}
\]
where $\tau_{\mathrm{crit}}(k, \hat{N}, \alpha)$ denotes the minimum empirical skill required to reject random guessing given $\hat{N}$ test samples (see Table~\ref{tab:acc-skill-grid} for detailed values). Intuitively, $\mathrm{LPC}(\pi_\theta, m) = 1$ as long as the model performs detectably above chance at depth $m$, even if the margin is small. 

We exclude $m=2$ from the analysis, as solving such graphs requires no lookahead: the model can simply return the node directly connected to $v_{\mathrm{target}}$, a shortcut achievable through shallow pattern-matching mechanisms such as induction heads \citep{olsson2022context}.

\paragraph{Strategy Discovery Depth Ceiling} To efficiently identify the maximum depth at which each model can discover a planning strategy, we adopt a progressive training procedure. Starting from $m = 3$, we train the model from its base initialization on $\mathcal{G}_{(k,m)}$ with $k \in \mathcal{K}$ and evaluate on the same configuration, checking whether $\mathrm{LPC}(\pi_\theta, m) = 1$. If so, we increment $m$ and repeat process on the new depth. The highest $m$ satisfying this criterion defines the strategy \emph{discovery depth ceiling} of the model. Each model is trained only on problems requiring a single planning depth, although LLMs are evaluated on multiple depths to test generalization. This ensures that success at greater depths cannot be explained by bootstrapping from simpler training instances. 

\subsection{Models}
\paragraph{Transformer} We begin by studying whether latent planning strategies can be discovered in the most controlled setting, where no prior planning knowledge is present in the model. To isolate strategy discovery from pre-training effects, we train an autoregressive transformer following the standard GPT-2 architecture \citep{radford2019language} from-scratch, using GELU \citep{hendrycks2016gaussian} as the activation function.
Specifically, it consists of 8 layers (1 head, 128 hidden dimension), yielding a total of 1.6M parameters.

\paragraph{Language models} We further evaluate and fine-tune large language models with different parameter scales.
For open-source models, this includes Qwen-2.5 (7B/32B) \citep{qwen2.5} and Qwen-3 (8B/32B) \citep{qwen3}.
We additionally include two proprietary models: GPT-4o\footnote{Specifically, \texttt{gpt-4o-2024-08-06} is used for both evaluation and fine-tuning via the OpenAI API.} \citep{openai2024gpt4technicalreport}, and GPT-5.4\footnote{\texttt{gpt-5.4-2026-03-05}. Fine-tuning is not publicly available for this model.} \citep{singh2025openai}. To ensure a fair comparison, we disable the hidden reasoning trace of GPT-5.4 by setting the reasoning effort parameter to none, so that the model must solve the task solely from a single forward pass without any intermediate deliberation. 

Full training details and hyperparameters are provided in Appendix~\ref{appendix:train}.

\subsection{Strategy Probing via Attention in Transformers}
Since the transformer is initialized without any pre-existing knowledge, its attention patterns reflect the unconfounded strategies acquired during training. To determine whether the model selectively attends to relevant graph components during inference, we define a \emph{backtracking ratio} (BR) that measures the fraction of edge-token attention allocated to edges on the path between $v_{\mathrm{target}}$ and $v_{\mathrm{source}}$ (formal definitions are provided in Appendix~\ref{appendix:attention_probing}). A near-uniform BR likely reflects parallel or blind search, whereas a BR that concentrates on path edges suggests that the model likely adopts a backtracking-based strategy.

%% file: secs/02_discussions.tex
\section{Results with Transformers Trained from Scratch}

\paragraph{\emph{Finding 1: Small Transformers Can Discover (Shallow) Latent Planning Strategies}}
As shown in Table~\ref{tab:acc_trained}, the transformer successfully solves most branch factors at depth $m=3$, while its skill rapidly decays to near or below zero as either the planning depth $m$ or branch factor $k$ increase. Unlike previous work reporting a complete failure to learn planning strategies under next-token prediction \citep{bachmann2024pitfalls}, our results show that such strategies can emerge, albeit within a narrow planning horizon.

\paragraph{\emph{Finding 2: Depth Imposes a Discovery Ceiling, Whereas Breadth Imposes a Capacity Limit}}
\begin{figure}[t]
    \centering
    \begin{minipage}{0.48\textwidth}
        \centering
        \vspace{-7mm}
        \includegraphics[width=\linewidth]{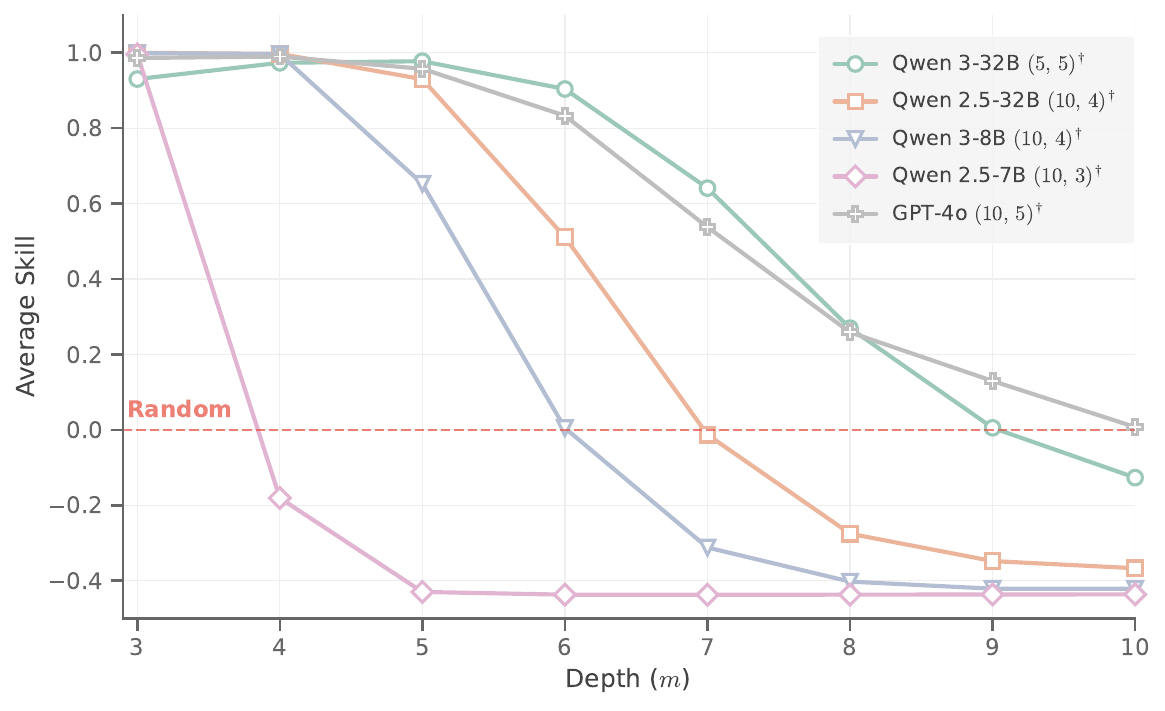}
        \vspace{-5mm}
        \caption{\textbf{Out-of-distribution (OOD) generalization of latent planning across depths.} Each model is trained on a \emph{single} configuration $(k^*, m^*)$ (marked with \textsuperscript{†}), and evaluated on unseen graphs at all depths \emph{without} further training. All models achieve near-perfect skill at depths up to $m^*$, and maintain high performance beyond the training depth, although this generalization systematically decays as the test depth increases.}
        \label{fig:ood_generalization}
    \end{minipage}%
    \hfill
    \begin{minipage}{0.48\textwidth}
        \centering
        \vspace{-2.6mm}
    \includegraphics[width=\linewidth]{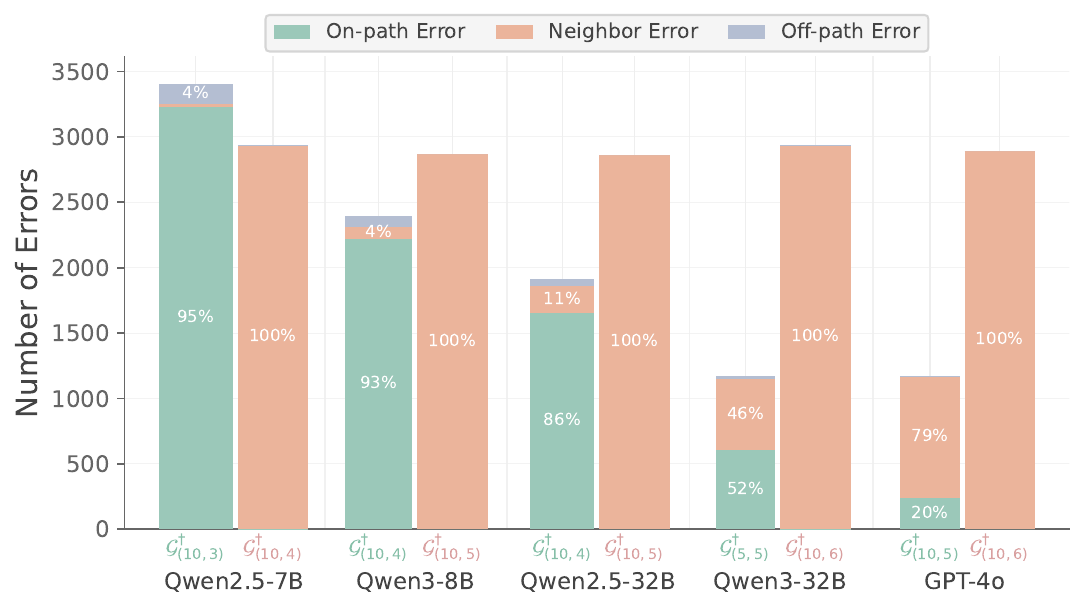}
        \vspace{-5mm}
        \caption{\textbf{Error distributions of fine-tuned LLMs.} Each model is trained on a \emph{single} configuration (marked with \textsuperscript{†}). \textcolor[RGB]{135,185,150}{$\mathcal{G}^{\dagger}$} denotes the most complex solvable configuration $(k^*, m^*)$; \textcolor{orange}{$\mathcal{G}^{\dagger}$} denotes $(10, m^*+1)$, where all models remain at random guessing. The high proportion of on-path errors indicates that models successfully identify the correct branch but fail to complete all planning steps, revealing a generalization ceiling.}
    \label{fig:on_path_ratio}
    \end{minipage}
\end{figure}
Notably, the model exhibits a qualitatively distinct failure mode at branch factor $k=10$: its empirical skill consistently remains below zero, unlike failures induced by increasing depth, where skill remains around zero. To better understand these observed failure cases, we further analyze the training dynamics. As shown in Figure~\ref{fig:transfomer_loss}, the model typically exhibits a two-stage learning process. In the first stage, it learns a simple heuristic: predicting any valid neighbor of the source node, which brings validation accuracy to the random baseline. In the second stage, the model attempts to learn the full multi-step planning strategy. Success leads to perfect validation accuracy, whereas failure results in the model simply memorizing the training graphs and overfitting.

An examination of these stages reveals a clear distinction between breadth and depth limitations.  We observe that an increase in the branch factor generally increases the number of training steps required to complete the first stage. When $k=10$, the model fails to learn even the basic heuristic, indicating a fundamental capacity limitation in processing complex local connectivity. In contrast, failures caused by increasing planning depth occur strictly during the second stage: the model successfully acquires the local heuristic but fails to discover the complete multi-step algorithm, representing a \textbf{discovery ceiling in planning depth}. Importantly, increasing model depth, hidden dimension, or the number of attention heads does not overcome this limitation (see Table~\ref{tab:ablation}).

\paragraph{\emph{Finding 3: Successful Latent Planning in Transformers Seems to Rely on Backtracking Strategies}}
\begin{wraptable}{R}{0.4\textwidth}
\vspace{-4.5mm}
\centering
\small
\renewcommand{\arraystretch}{1}
\setlength{\tabcolsep}{4pt}
\caption{
Average backtracking ratio (BR) of the trained transformer on the test set. 
\textbf{Blue-highlighted bold text} indicates configurations where the model successfully learned a latent planning strategy, while \textcolor{failtext}{gray text} indicates failure.
}
\vspace{0mm}
\label{tab:br-transformer}
\begin{tabular}{l ccccc}
\toprule
\multirow{2}{*}{Depth} & \multicolumn{5}{c}{Branch Factor ($k$)} \\
\cmidrule(lr){2-6}
 & 2 & 3 & 4 & 5 & 10 \\
\midrule
$m=3$ & \win{0.50} & \win{0.51} & \win{0.78} & \win{0.73} & \fail{0.10} \\
$m=4$ & \fail{0.49} & \fail{0.34} & \fail{0.25} & \fail{0.20} & \fail{0.10} \\
\bottomrule
\end{tabular}
\end{wraptable}
To probe the implicit planning strategy, we measure the average BR on the test set for each $\mathcal{G}_{(k,m)}$, with planning depths $m \in \{3,4\}$ and branch factors $k \in \mathcal{K}$. As shown in Table~\ref{tab:br-transformer}, the BR is close to $\frac{1}{k}$ for all configurations that the model fails to solve, indicating that attention over the graph is approximately uniform, consistent with a parallel or blind search pattern. In contrast, for configurations that the model successfully solves with $k \geq 3$, the BR is substantially higher than the uniform baseline, suggesting that the model has discovered a backtracking-based strategy during training. Interestingly, when $k=2$, both the unsuccessful configuration (with $m=4$) and the successful configuration (with $m=3$) exhibit a BR close to $\frac{1}{2}$. We attribute it to the unique topology of this graph: the model can perform a forward traversal by randomly selecting a neighbor; if it fails to reach $v_{\mathrm{target}}$, the model can select the alternative neighbor through elimination, thereby avoiding the need for backtracking.

We further conduct a qualitative analysis by visualizing the attention maps of the trained transformer (see Appendix~\ref{appendix:attention_visualization}). For successful configurations, as the branch factor increases, the attention of the model becomes increasingly concentrated on nodes along the path from $v_{\mathrm{target}}$ to $v_{\mathrm{source}}$, following a clear backward order from deeper to shallower layers. In contrast, for failed configurations, no discernible pattern emerges in the attention maps.

\section{Results with LLMs}

\paragraph{\emph{Finding 4: Pre-Trained LLMs Exhibit Shallow Latent Planning That Scaling Alone Does Not Resolve}}
We evaluate LLMs under zero-shot and few-shot prompting without any task-specific training (detailed in Appendix~\ref{appendix:zero_few_shot}). As shown in Figure~\ref{fig:lpc_bar} and Table~\ref{tab:acc_untrained}, all models exhibit latent planning ability at shallow depths, with LPC ranging from $3$ (Qwen 3-8B) to $5$ (GPT-5.4) in the zero-shot setting. Few-shot prompting provides consistent gains, extending the LPC of GPT-5.4 to $7$ (Figure~\ref{fig:lpc_bar} and Table~\ref{tab:gpt54_skill_m78}), yet all models eventually fail as planning depth increases. Notably, even GPT-5.4, which represents a significant leap in general capability over GPT-4o \citep{zheng2023judging}, extends the zero-shot planning depth by only one step, and the depth barrier persists even with in-context demonstrations. Furthermore, this limitation cannot be attributed to graph size: small but deep graphs (e.g., $\mathcal{G}_{(2,6)}$ with only 13 nodes) remain unsolvable while larger but shallower graphs are solved easily, confirming that planning depth is the fundamental bottleneck.

\paragraph{\emph{Finding 5: Scaling Improves Breadth but Struggles to Overcome the Discovery Bottleneck}} We further fine-tune the pre-trained LLMs to evaluate their capacity for latent strategy discovery. 
The evaluation results for the fine-tuned models are detailed in Table~\ref{tab:acc_trained}.  Comparing with the from-scratch transformer reveals that scaling primarily improves the ability to handle greater planning breadth rather than to discover deeper planning strategies. For instance, while the from-scratch transformer fails entirely on the highly branched configuration with $m=3$ and $k=10$, Qwen 2.5-7B perfectly solves this task. However, this advantage in breadth does not extend to deeper planning. At $m=4$, Qwen 2.5-7B fails across all branch factors, indicating that its discovery depth ceiling remains the same as that of the from-scratch model despite the much larger model scale. Further scaling (up to 32B or even to massive frontier models) yields only marginal improvements in planning depth. Specifically, Qwen 3-32B and GPT-4o successfully solve configurations up to $m=5$ but fail completely at $m=6$, where none of the evaluated models can significantly reject the random-guessing hypothesis, highlighting the persistence of the discovery bottleneck.

Surprisingly, the training dynamics of these fine-tuned LLMs (detailed in Appendix~\ref{appendix:llm_training}) closely mirror the two-stage learning process observed in the from-scratch transformer. Unlike the small transformer, which also fails in the first stage at high branch factors, LLM failures occur exclusively in the second stage, confirming that the bottleneck lies in strategy discovery rather than a mere parameter deficit.

\paragraph{\emph{Finding 6: Discovered Strategies Generalize Beyond Training Horizons}}
We further evaluate models' out-of-distribution (OOD) length generalization. For each base model, we select the checkpoint fine-tuned \emph{only} on its most complex solvable configuration $(k^*, m^*)$, where 
\[(k^*, m^*) = \arg\max_{\mathrm{lex}(m,\,k)} \bigl\{ (k, m)\, \big\vert\, \text{Skill}(\pi_\theta, k, m) \approx 1 \bigr\}. \]
We then evaluate this checkpoint on all combinations of planning depths $m \in \{3, 4, \dots, 10\}$ and branch factors $k \in \mathcal{K}$, computing the average skill across branch factors at each depth.

As illustrated in Figure~\ref{fig:ood_generalization}, all evaluated models exhibit strong interpolation, achieving near-perfect skills on unseen graphs  with $m \leq m^*$. Furthermore, with the exception of Qwen 2.5-7B, these models successfully extrapolate, performing significantly above the random baseline at depth $m^*+1$. This extrapolation capacity also scales with model capability: Qwen 3-32B maintains above-chance performance up to $m^*+3$, while GPT-4o extends this to $m^*+4$. This robust OOD performance confirms that models successfully internalize a generalizable path-finding strategy, and that the primary bottleneck lies in discovering the strategy under sparse, final-answer supervision.  Once discovered, models can effectively execute the learned strategy beyond their training horizons---even to planning depths they are not able to learn when trained on directly.

\paragraph{\emph{Finding 7: On-Path Failures Reveal a Generalization Depth Ceiling}}
As planning depth increases, we observe that several models exhibit skills that fall well below the random baseline, indicating that these errors are not mere random guesses, but rather the result of a systematic yet flawed execution strategy. To investigate this phenomenon, we conduct a fine-grained structural analysis of the error samples. For each incorrect prediction $\hat{v}$, we compute its shortest-path distance to the source node $v_{\text{source}}$ and the target node $v_{\text{target}}$, formally denoted as: $d_1 = d(\hat{v}, v_{\text{source}}), \quad d_2 = d(\hat{v}, v_{\text{target}})$.

For a specific graph configuration $\mathcal{G}_{(k,m)}$, we partition the errors into three mutually exclusive categories:
\begin{itemize}
    \item \textbf{1-hop neighbor errors} ($\mathcal{E}_{\text{1-hop}}$): Defined by $\{ \hat{v} \mid d_1 = 1 \}$, indicating predictions of a direct but wrong neighbor of the source.
    \item \textbf{On-path errors} ($\mathcal{E}_{\text{on-path}}$): Defined by $\{ \hat{v} \mid d_1 \neq 1 \land d_2 < k - 1 \}$, indicating the model successfully selects the correct branch but outputs an incorrect hop.
    \item \textbf{Off-path errors} ($\mathcal{E}_{\text{off-path}}$): Defined by $\{ \hat{v} \mid d_1 \neq 1 \land d_2 \geq k \}$, indicating the prediction lands on an incorrect branch or the source node itself.
\end{itemize}

To ensure a fair comparison, we restrict the analysis to the overlapping subset of test data between open-source and closed-source models. As a baseline, we additionally evaluate checkpoints trained on $\mathcal{G}_{(10, m^* + 1)}$, where all models remain in the first training stage and produce random guesses. As illustrated in Figure~\ref{fig:on_path_ratio}, all evaluated models exhibit a substantial proportion of on-path errors, a pattern that never emerges in models that remain in the first training stage. Notably, for Qwen 2.5-7B, on-path errors constitute $95\%$ of all errors, demonstrating that the model consistently attempts to execute the learned strategy and successfully locates the correct branch, but fails to complete all planning steps. This reveals a generalization depth ceiling: even when a model internalizes the correct algorithm, its ability to execute this strategy latently over extended horizons remains strictly bounded.

Furthermore, we observe an inverse relationship between the capability of the model and the ratio of on-path errors. Here, we hypothesize that models with larger capacities learn multiple planning strategies simultaneously, whereas smaller models strictly adhere to a single dominant strategy due to capacity limitations.

Together, Findings 6 and 7 suggest that the strategies discovered through fine-tuning can be integrated with structural reasoning capabilities (as ``world modeling'') acquired during pre-training, enabling models to generalize well beyond their training horizon.

\paragraph{\emph{Finding 8: Dense Supervision Bypasses the Discovery Bottleneck}}

The discovery bottleneck identified above is specific to sparse, final-answer supervision. When models externalize their reasoning through an explicit backtracking chain of thought, they solve graphs requiring twenty lookahead steps with minimal effort, converging in only $20$ training updates (Figure~\ref{fig:cot_loss} in Appendix~\ref{appendix:cot}). This confirms that the task itself is not inherently difficult, and the discovery bottleneck can be bypassed easily by explicitly teaching the planning strategy through dense supervision. 
We further show that, using the ICoT framework \citep{deng2023implicit,deng2024explicit}, the explicitly learned strategy can be progressively compressed back into implicit reasoning, enabling the small transformer to achieve perfect latent planning up to $m=6$ at $k=2$. However, this distillation remains limited by the representational capacity of the model and does not fully scale to more complex graph configurations (Appendix~\ref{appendix:icot}).